\documentclass[lettersize,journal]{IEEEtran}

\usepackage{amsmath,amsfonts}
\usepackage{algorithmic}
\usepackage[ruled]{algorithm2e}
\usepackage{subcaption}
\usepackage{array}
\usepackage{textcomp}
\usepackage{multicol}
\usepackage{stfloats}
\usepackage{url}
\usepackage{verbatim}
\usepackage{graphicx}
\usepackage{cite}
\usepackage{diagbox}
\usepackage{pifont}
\usepackage{amssymb}
\usepackage{multirow}
\usepackage{xcolor}
\usepackage{bm}
\usepackage{makecell}
\usepackage[normalem]{ulem}

\newcolumntype{C}[1]{>{\centering\let\newline\\\arraybackslash\hspace{0pt}}m{#1}}

\newcommand{\bI}{\mathbf{I}}

\newcommand{\bzero}{\mathbf{0}}

\newcommand{\bz}{\mathbf{z}}

\newcommand{\bepsilon}{{\boldsymbol{\epsilon}}}

\newcolumntype{A}{>{\columncolor{green!10}}}
\usepackage[utf8]{inputenc}
\usepackage{hyperref}
\begin{document}

\title{INDIGO+: A Unified INN-Guided Probabilistic Diffusion Algorithm for Blind and Non-Blind Image Restoration}

\author{Di You,~\IEEEmembership{Student Member,~IEEE,} and Pier Luigi Dragotti,~\IEEEmembership{Fellow,~IEEE}
}

\markboth{Journal of \LaTeX\ Class Files,~Vol.~14, No.~8, August~2021}%
{Shell \MakeLowercase{\textit{et al.}}: A Sample Article Using IEEEtran.cls for IEEE Journals}


\maketitle

\begin{abstract}
Generative diffusion models are becoming one of the most popular prior in image restoration (IR) tasks due to their remarkable ability to generate realistic natural images. Despite achieving satisfactory results, IR methods based on diffusion models present several limitations. First of all, most non-blind approaches require an analytical expression of the degradation model to guide the sampling process. Secondly, most existing blind approaches rely on families of pre-defined degradation models for training their deep networks. The above issues limit the flexibility of these approaches and so their ability to handle real-world degradation tasks.

In this paper, we propose a novel INN-guided probabilistic diffusion algorithm for non-blind and blind image restoration, namely INDIGO and BlindINDIGO, which combines the merits of the perfect reconstruction property of invertible neural networks (INN) with the strong generative capabilities of pre-trained diffusion models. Specifically, we train the forward process of the INN to simulate an arbitrary degradation process and use the inverse to obtain an intermediate image that we use to guide the reverse diffusion sampling process through a gradient step. 
{We also introduce an initialization strategy, to further improve the performance and inference speed of our algorithm.}
Experiments demonstrate that our algorithm obtains competitive results compared with recently leading methods both quantitatively and visually on synthetic and real-world low-quality images.
\end{abstract}


\begin{figure*}[!tp]\footnotesize 
	\centering
\hspace{-0.2cm}
\begin{tabular}{c@{\extracolsep{0.1em}}c@{\extracolsep{0.1em}}c@{\extracolsep{0.1em}}c@{\extracolsep{0.1em}}c@{\extracolsep{0.1em}}c@{\extracolsep{0.1em}}c@{\extracolsep{0.1em}}c@{\extracolsep{0.1em}}c}
     &\includegraphics[width=0.15\textwidth]{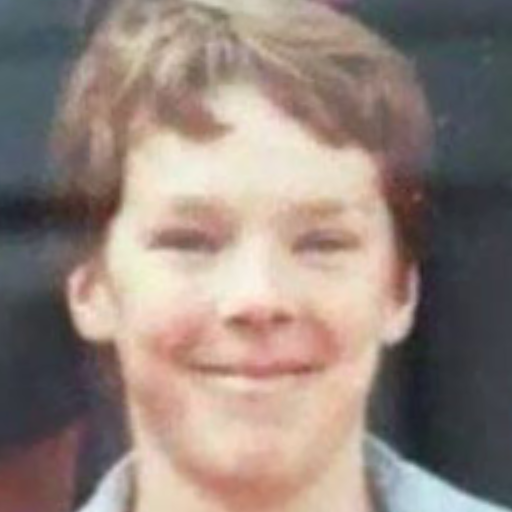}~
     &\includegraphics[width=0.15\textwidth]{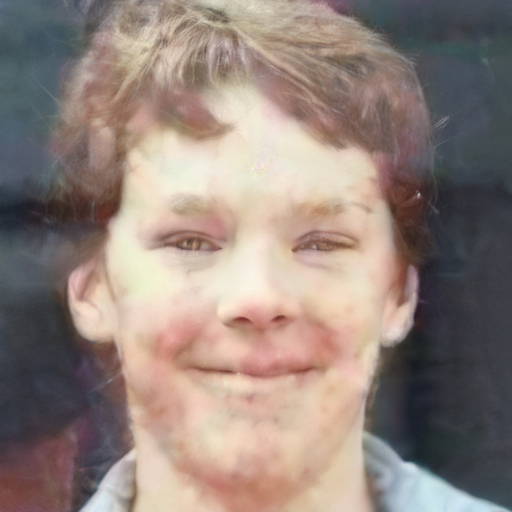}~
		&\includegraphics[width=0.15\textwidth]{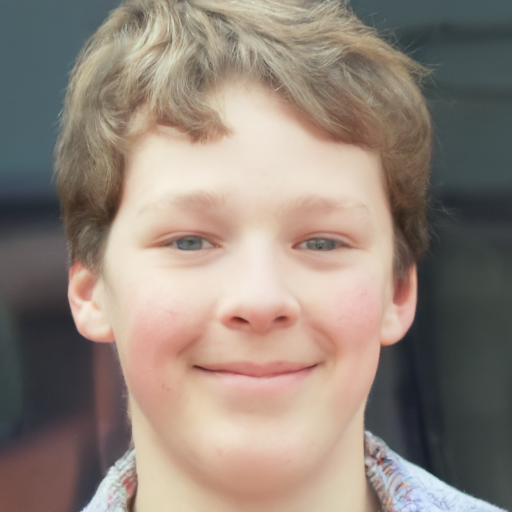}~
   &\includegraphics[width=0.15\textwidth]{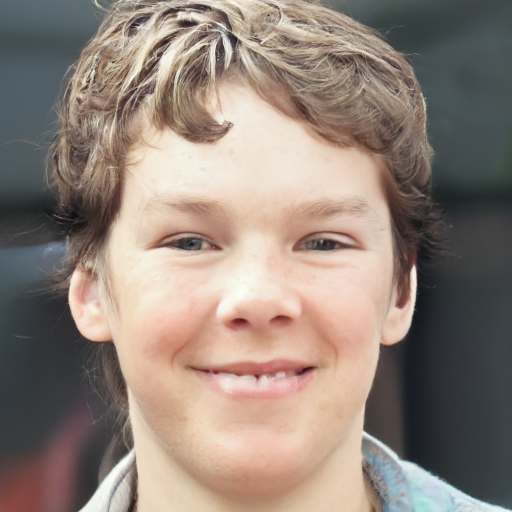}~&\includegraphics[width=0.15\textwidth]{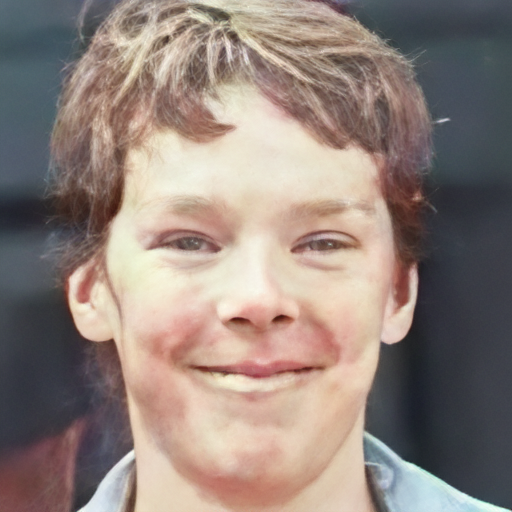} ~
		&\includegraphics[width=0.15\textwidth]{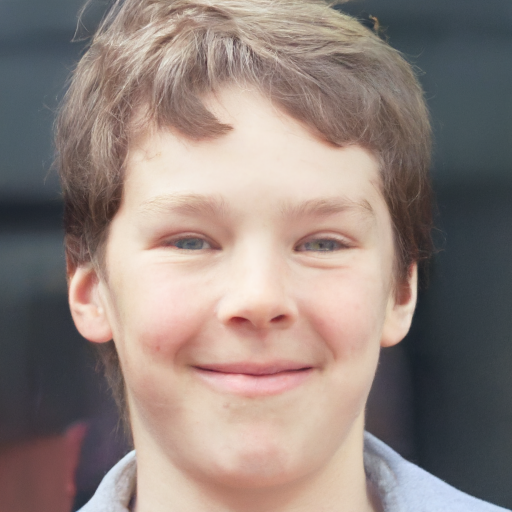}~
		\\
  & (a) Input &(b) DR2 \cite{wang2023dr2} &(c) DifFace \cite{yue2022difface} &(d) PGDiff \cite{yang2023pgdiff} &(e) {StableSR} \cite{stablesr} &(f) \textbf{{Ours}} \\
	\end{tabular}
		\caption{{Comparisons with state-of-the-art blind image restoration approaches  \cite{wang2023dr2,yue2022difface,yang2023pgdiff,stablesr} on the real-world low-quality images. Our algorithm produces high-quality reconstruction results and preserves more details than the recent leading methods. (Zoom in for best view).}} 
	\label{fig:1}
\end{figure*}
\section{Introduction}
\IEEEPARstart{I}{n} this paper, we explore a new way to employ diffusion models for image restoration. Image restoration (IR) is a typical inverse problem aiming to recover high-quality images from their noisy and degraded measurements. 
In a typical restoration problem, one observes $\bm{y} =\mathcal{H}(\bm{x},\bm{n})$, where $\bm{y}$ is the degraded and noisy version of the original image $\bm{x}$ and $\bm{n}$ is some noise. The degradation process $\mathcal{H}$ can be linear or non-linear and is often unknown. {In this paper, we classify these inverse problems as blind or non-blind IR problems based on whether we can access labeled training pairs that are degraded in the same way as the measurement (fully supervised setting) or not. 
Specifically, the non-blind case includes two situations: either we know the expression of the forward operator, or we can simulate it without knowing its analytical expression and can therefore produce labeled image pairs. In the blind case, we do not have access to the labeled data and therefore do not have any way to simulate the actual degradation process.}

This inverse problem is normally solved by addressing the classic trade-off between a data
fidelity and a regularization term based on proper priors. While this approach goes back to the classic Tikhonov
regularization, a wealth of new models and regularizers have emerged over the years often driven by the idea
of sparsity. These have led to the development of many model-based reconstruction methods. 

Recently, there has been a shift towards developing data-driven approaches, in particular based on deep
learning architectures where the regularization is implicitly learned through the data and we refer to \cite{review} for a recent
review on the topic. The plug-and-play (PnP)  framework \cite{a} is a typical example where the prior can be learned through data. PnP is based on iterating between a step that enforces some forms of data consistency and a denoising step where the denoiser can be implemented with a deep neural network. In this context, the denoiser effectively acts as a data-driven regularizer. Interestingly, this heuristic has led to many remarkable results and we refer to \cite{b} for a recent overview on the topic.

The generative prior of diffusion models \cite{ddpm,ddim,dhariwal2021diffusion,song2019generative,songscore} has now become one of the most popular priors in image restoration problems due to their remarkable ability to approximate the natural image manifold. A line of work \cite{ilvr,ccdf,Repaint,kadkhodaie2020solving,snips,DDRM,DDNM,songmedical,pseudoinverse,dps,dolce,you2023indigo} has focused on leveraging the rich image priors and strong generative capability of pretrained diffusion models to solve IR problems. Among them, earlier works \cite{ilvr,ccdf,Repaint,kadkhodaie2020solving} have 
focused on linear degradation models and noiseless measurements.
Two popular categories of approaches have then been proposed for investigating noisy and non-linear inverse problems. Decomposition-based approaches \cite{snips,DDRM,DDNM,songmedical} run singular value decomposition (SVD), range-null space decomposition or matrix decomposition on intermediate results during iterations to guide the sampling process. Gradient-based approaches \cite{pseudoinverse,dps,dolce,you2023indigo} 
propose to incorporate consistency-imposing gradient steps in between the reverse diffusion steps.

Despite achieving satisfactory results, the aforementioned methods have inevitably limited generalization capabilities because their algorithms are designed under non-blind degradation settings. To alleviate this limitation, several diffusion model-based approaches \cite{blinddps,gibbsddrm,gdp} have been recently developed for unknown degradation operators in specific tasks such as image deblurring \cite{blinddps,gibbsddrm} and low-light enhancement \cite{gdp}.
Towards a more diverse and complicated degradation process,
DifFace~\cite{yue2022difface} introduces a pre-trained IR network $g(\cdot)$ (e.g., based on CNN or Transformer) to obtain an initial distortion-invariant clean image as a starting point, $x_N$, for the subsequent diffusion sampling process.
In the framework of DR2~\cite{wang2023dr2}, 
smooth results are first predicted by an iterative refinement similar to ILVR \cite{ilvr} during sampling
and then further processed by a pre-trained IR network $g(\cdot)$ to achieve high-quality details. 
PGDiff \cite{yang2023pgdiff} constrains the high-quality image space during the posterior sampling process 
with a constraint on the MSE between $ g(\bm{{y}})$ and the denoised intermediate output obtained at step $t$ of the diffusion reverse process.
StableSR  \cite{stablesr} uses the Stable Diffusion model \cite{stablediffusion} for image super-resolution and is further equipped with a time-aware encoder and a controllable feature wrapping module.

The above non-blind and blind IR approaches 
have demonstrated the effectiveness of the generative diffusion models for IR tasks. However, they are faced with the following limitations: 
(1) In the task of non-blind IR, most existing approaches require a closed-form expression of the degradation model to guide the sampling process. However, the image processing pipeline of many modern imaging systems is so complex that it is often impossible to describe it explicitly. (2) In the task of blind IR, most existing blind IR approaches rely on pre-defined degradation models for training the IR network $g(\cdot)$, which also limits their flexibility in real-world scenarios.

To address the above issues, we propose an INN-guided probabilistic diffusion algorithm for both non-blind~\footnote{The work on non-blind inverse problem was presented in part at IEEE MMSP conference 2023 \cite{you2023indigo}.} and blind image restoration. During the sampling process of diffusion model, we impose an additional data-consistency step by introducing an off-the-shelf light-weight invertible neural network (INN).
Specifically, we pre-train the forward process of INN to simulate an arbitrary degradation process. At testing stage we alternate between an unconditional diffusion sampling step that gives us an intermediate image consistent with the diffusion model and a consistency step guided by the INN that forces the reconstruction to be consistent with the measurements. In particular, given at each step an estimated image, the forward part of the INN produces a coarse image which we then force to be consistent with the measurements and the details estimated by the diffusion process. We then use the inverse part of the INN as a reconstruction process to obtain an intermediate result that guides the next step of the reverse diffusion process.
Therefore, our method guides the sampling towards satisfying the consistency constraint while maintaining rich details provided by the diffusion prior. In the task of non-blind IR, INN is pretrained with datasets on any specific degradation, so it is no longer limited by the requirement of knowing the analytical expression of the degradation model. In the task of blind IR, we first initialize the parameters of INN by training it with synthetic dataset pairs that model different degradation processes. Then, by alternating between refining the INN parameters for the unknown degradation model and updating intermediate image results with the guidance of INN during sampling, our approach is more flexible and can handle unknown degradation settings in real-world scenarios.

We summarize our contributions as follows:
\begin{itemize}
\item We propose a novel INN-guided probabilistic diffusion algorithm for non-blind and blind image restoration, namely INDIGO and BlindINDIGO.
In contrast to most existing approaches, our algorithm introduces prior degradation information to the diffusion reverse process by simulating it with INN, which help to boost IR performance and improve flexibility.
\item To the best of our knowledge, this is the first attempt to combine the merits of the perfect reconstruction property of INN with strong generative prior of diffusion models for blind image restoration. With the help of INN, our algorithm effectively estimates the details lost in the degradation process and is able to handle arbitrary degradation processes.
\item 
{We further introduce an initialization strategy to accelerate our algorithm by reducing the number of timestep.}
\item Extensive experiments show that our approach for both non-blind and blind image restoration achieves state-of-the-art results compared with other methods on synthetically degraded and real low-quality images (see 
Fig.~\ref{fig:1} for an example).
\end{itemize}

\section{Background}
\subsection{Review of Denoising Diffusion Probabilistic Models}
Diffusion models, e.g. \cite{ddpm,ddim,dhariwal2021diffusion,song2019generative,songscore,lu2022dpm,ho2022classifier}, sequentially corrupt training data with slowly increasing noise, and then learn to reverse this corruption in order to form a generative model of the data. Here we describe a classic diffusion model: denoising diffusion probabilistic model (DDPM)~\cite{ddpm}.
DDPM defines a $T$-step forward process transforming complex data distribution into simple Gaussian noise distribution and a $T$-step reverse process recovering data from noise. 
The forward process slowly adds random noise to data, where, in the typical setting, the added noise has a Gaussian distribution.
Consequently, the forward process yields the present state $\bm{x}_{t}$ from the previous state $\bm{x}_{t-1}$:
\begin{equation}
    q(\bm{x}_{t}|\bm{x}_{t-1})=\mathcal{N}(\bm{x}_{t};\sqrt{1-\beta_{t}}\bm{x}_{t-1},\beta_{t}\mathbf{I})\quad
    \label{eq:ddpm forward 1}
\end{equation}
where $\bm{x}_{t}$ is the noisy image at time-step $t$, $\beta_{t}$ is a predefined scale factor. As noted in \cite{ddpm}, the above process allows us to sample an arbitrary state $\bm{x}_{t}$ directly from the input $\bm{x}_{0}$ as follows:
\begin{equation}
\bm{x}_{t}=\sqrt{\Bar{\alpha}_{t}}\bm{x}_{0}+ \sqrt{1-\Bar{\alpha}_{t}} \boldsymbol{\epsilon} 
    \label{eq:xt}
\end{equation}
where $\alpha_{t} = 1- \beta_{t}$, $\quad \Bar{\alpha}_{t} = \prod_{i=0}^{t}\alpha_{i}$ and $\boldsymbol{\epsilon}\sim \mathcal{N}(0,\mathbf{I})$.
For the reverse process, we can calculate the posterior distribution $q(\bm{x}_{t-1}|\bm{x}_{t},\bm{x}_{0})$ using Bayes theorem and write the expression of $\bm{x}_{t-1}$ using Eq.~(\ref{eq:xt}) as follows:
\begin{equation}
\begin{split}
   \bm{x}_{t-1} = \frac{1}{\sqrt{\alpha_{t}}}	\left( \bm{x}_{t} - \frac{1-\alpha_{t}}{\sqrt{1-\Bar{\alpha}_{t}}} \boldsymbol{\epsilon} \right) + \sigma_t\mathbf{z},
    \label{eq:ddpm reverse 1}
\end{split}
\end{equation}
where $\mathbf{\sigma}_{t}$$=$$\sqrt{\frac{1-\Bar{\alpha}_{t-1}}{1-\Bar{\alpha}_{t}}\beta_{t}}$ and $\mathbf{z}\sim \mathcal{N}(0,\mathbf{I})$.
To predict the noise $\boldsymbol{\epsilon}$ in the above equation, DDPM uses a neural network  $\boldsymbol{\epsilon}_{\boldsymbol{\theta}}(\bm{x}_{t},t)$ for each time-step $t$.
To train  $\boldsymbol{\epsilon}_{\boldsymbol{\theta}}(\bm{x}_{t},t)$, DDPM {\color{black}{uniformly samples a $t$ from $ \{1,...,T\}$}} and updates the network parameters $\boldsymbol{\theta}$ with the following gradient descent step:
\begin{equation}
    \nabla_{\boldsymbol{\theta}}||\boldsymbol{\epsilon}-\boldsymbol{\epsilon}_{\boldsymbol{\theta}}(\sqrt{\Bar{\alpha}_{t}}\bm{x}_{0} + \sqrt{1-\Bar{\alpha}_{t}}\boldsymbol{\epsilon},t)||^{2}_{2},
    \label{eq:7}
\end{equation}
where $\bm{x}_{0}$ is a clean image from the dataset and $\boldsymbol{\epsilon}\sim \mathcal{N}(0,\mathbf{I})$ is random noise. 
By replacing $\boldsymbol{\epsilon}$ with the approximator $\boldsymbol{\epsilon}_{\boldsymbol{\theta}}(\bm{x}_{t},t)$ in Eq.~(\ref{eq:ddpm reverse 1}) and iterating it $T$ times,  DDPM can yield clean images $\bm{x}_{0}\sim q(\bm{x})$ from initial random noises $\bm{x}_{T}\sim\mathcal{N}(\mathbf{0},\mathbf{I})$, where $q(\bm{x})$ represents the image distribution in the training dataset.

Solvers of inverse problems that use diffusion models 
have shown remarkable performance and versatility, and can be divided into two groups. The first group of methods
\cite{SR3,saharia2022palette,SR3+,li2022srdiff,qiu2023diffbfr} has focused on designing and training conditional diffusion models suitable for image reconstruction tasks. The second group \cite{ilvr,ccdf,Repaint,kadkhodaie2020solving,snips,DDRM,DDNM,songmedical,pseudoinverse,dps,dolce,you2023indigo} has instead focused on keeping the training of unconditional diffusion models unaltered, and only modify the inference procedure to enable sampling from a conditional distribution. 
The approach proposed in this paper falls in the latter category and has the advantage of leveraging the pre-trained diffusion models to make them serve as a strong generative prior without the need of retraining diffusion models.

\begin{figure}
     \centering
     \begin{subfigure}[b]{0.5\textwidth}
         \centering
         \includegraphics[width=\textwidth]{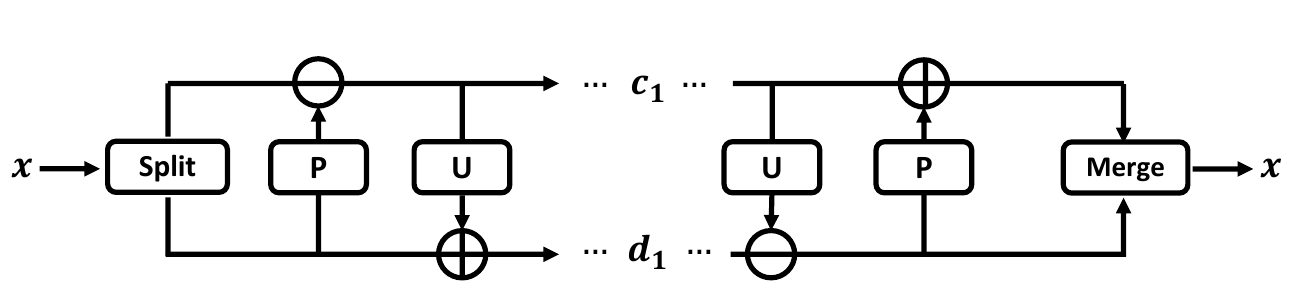}
         \caption{1-level lifting scheme}
         \label{fig:y equals x}
     \end{subfigure}
     \hfill
     \begin{subfigure}[b]{0.5\textwidth}
         \centering
         \includegraphics[width=\textwidth]{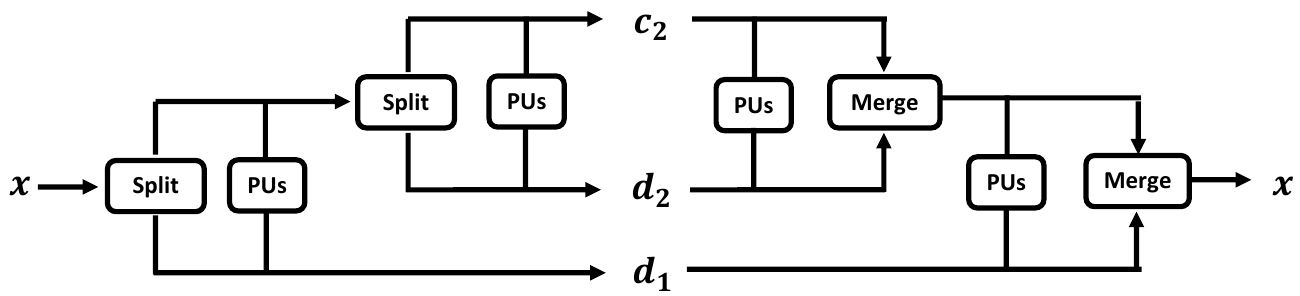}
         \caption{2-level lifting scheme}
         \label{fig:three sin x}
     \end{subfigure}
        \caption{The wavelet transform obtained using the lifting scheme.}
        \label{fig:lifting}
\end{figure}

\subsection{Wavelet Transform 
and Invertible Neural Networks}
\label{background_inn}
The wavelet transform is widely used in many imaging applications due to its ability to concentrate image features in a few large-magnitude wavelet coefficients, while small-value wavelet coefficients typically contain noise and can be shrunk or removed without affecting the image quality.
The lifting scheme \cite{daubechies1998factoring} is often used to construct a wavelet transform. As shown in Fig.~\ref{fig:lifting}(a), the forward wavelet transform converts the input signal into coarse and detail components and then the original signal is reconstructed by the inverse transform. Specifically, the lifting scheme first splits the signal $\bm{x} =({x}_{k})_{k \in Z}$ into an even $\bm{x}_e =({x}_{2k})_{k \in Z}$ and an odd part $\bm{x}_{o} =({x}_{2k+1})_{k \in Z}$. A predictor is used to predict the odd part from the even part, and thus the difference between the odd part and its prediction reflects high-frequency details $\bm{d}$ of the signal. Based on this difference, the update step is used to adjust the even part to make it a smoother coarse version $\bm{c}$ of the original signal. The above lifting procedure implementing the forward wavelet transform can be described as:
\begin{equation}
\begin{gathered}
\bm{d} = \bm{x}_{o} - P(\bm{x}_{e}) , \quad
\bm{c} = \bm{x}_{e}  + U(\bm{d}) .
\end{gathered}
\label{eq: forward}
\end{equation}
The inverse transform can immediately be found by reversing the operations and flipping the signs. Therefore, the original signal can be recovered as follows:
\begin{equation}
\begin{gathered}
\bm{x}_{e} = \bm{c} - U(\bm{d}), \quad
\bm{x}_{o} = \bm{d} + P(\bm{x}_{e}) .
\end{gathered}
\label{eq: backward}
\end{equation}
The above equations illustrate that no matter how $P$ and $U$ are chosen, the scheme is always invertible and thus leads to critically sampled perfect reconstruction filter banks  \cite{daubechies1998factoring}. Furthermore, this scheme allows multiple levels and multiple pairs of predictors and updates (see Fig.~\ref{fig:lifting}(b)).

Inspired by the above idea, Huang et al. \cite{winnet} propose a lifting-inspired invertible neural network (LINN) for image denoising. The forward transform of LINN non-linearly converts the input noisy image into coarse channel and detail channels. A denoising network performs the denoising operation on the detail part, and then the backward transform of the LINN reconstructs the denoised image using the original coarse channel and the denoised detail channels. In this architecture, INN consists of several invertible blocks where $P$ and $U$ in Eq.~(\ref{eq: forward}) and Eq.~(\ref{eq: backward}) become functions parameterized by neural networks. Specifically, the Predict and Update networks are applied alternatively to update the coarse and detail parts. The $m$-th pair of update and predict operations of the $k$-th level INN can be expressed as:
\begin{align} 
    \bm{d}^{k}_m &=  \bm{d}^{k}_{m-1} - P_m^k \left(\bm{c}^{k}_{m-1} \right), \\ 
    \bm{c}^{k}_m &=  \bm{c}^{k}_{m-1} + U_m^k \left(\bm{d}^{k}_m \right),
    \label{eq:LINNforward}
\end{align}
where $\bm{d}^{k}_m$ and $\bm{c}^{k}_m$ denotes the updated detail part and coarse part using the $m$-th Predict network $P_m^k(\cdot)$ and Update network $U_m^k(\cdot)$, respectively. 
Similarly, the inverse transform of the $k$-th level INN can be expressed as:
\begin{align} 
    \bm{c}^{k}_{m-1} &=  \bm{c}^{k}_m - U_m^k \left(\bm{d}^{k}_m \right),\\
    \bm{d}^{k}_{m-1} &=  \bm{d}^{k}_m + P_m^k \left(\bm{c}^{k}_{m-1} \right).
    \label{eq:LINNbackward}
\end{align}
There are also other choices for INN architectures, including coupling layer \cite{dinh2014nice}, affine coupling layer \cite{dinh2016density},  reversible residual network \cite{gomez2017reversible} and i-RevNet architecture \cite{jacobsen2018revnet}. The invertible architecture that we design in this paper is based on the lifting-inspired invertible blocks in \cite{winnet}. However, we use an alternative training strategy where we try to ensure that the coarse version produced by the network is as close as possible to the measured degraded image $\bm{y}.$

\section{INDIGO+ Approach}

\subsection{Overview}
\begin{figure}[t]
    \centering
    \includegraphics[width=0.5\textwidth]{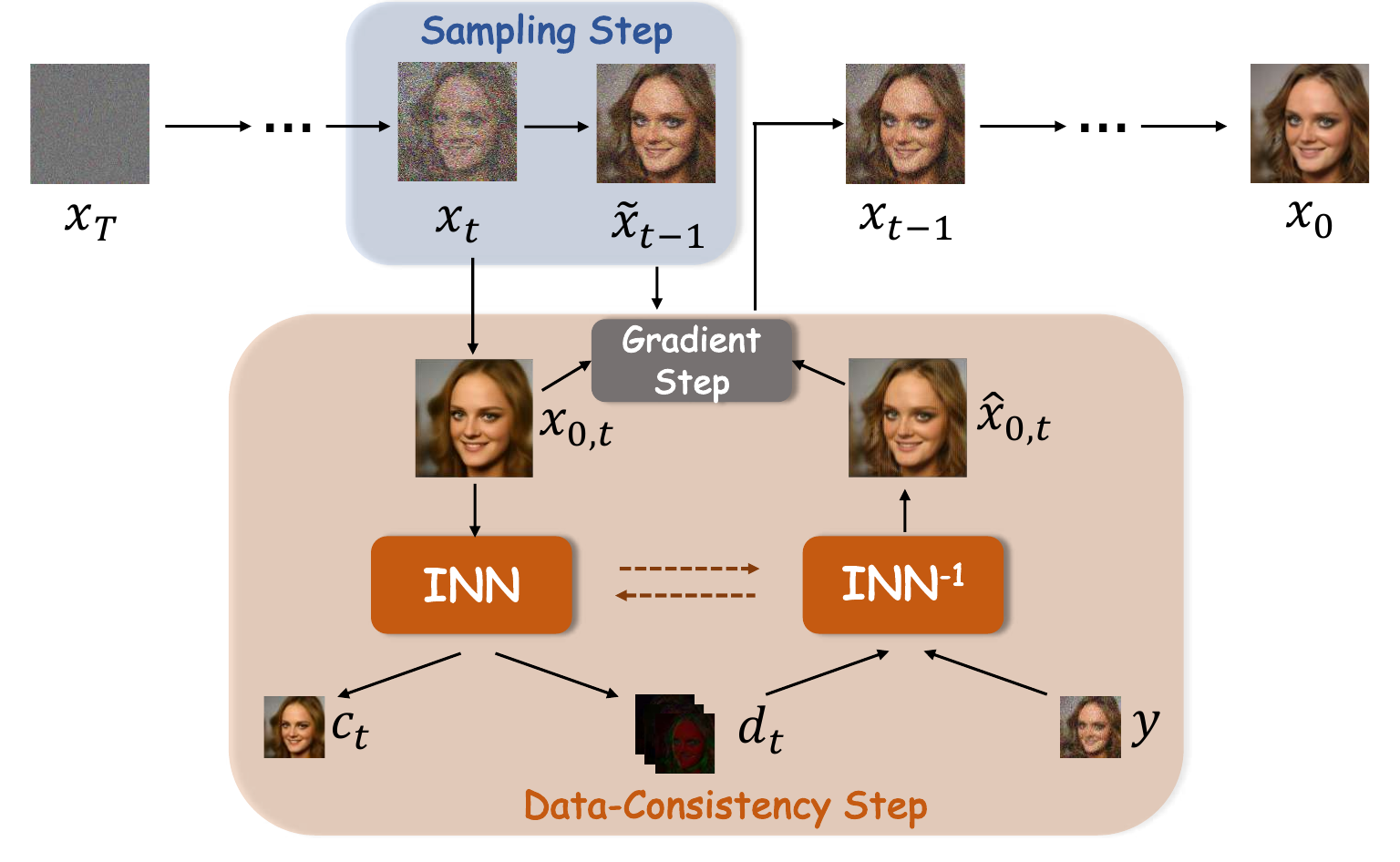}
\caption{ Overview of our INDIGO for non-blind image restoration. Given a degraded image $\bm{y}$ during inference, the diffusion posterior sampling is guided by our data-consistency step with INN at each step $t$. We show the detailed algorithm in Algorithm~\ref{algo:nonblind}.  }
    \label{fig:non-blind}
\end{figure}

\begin{figure*}[t]
    \centering
    \includegraphics[width=0.7\textwidth]{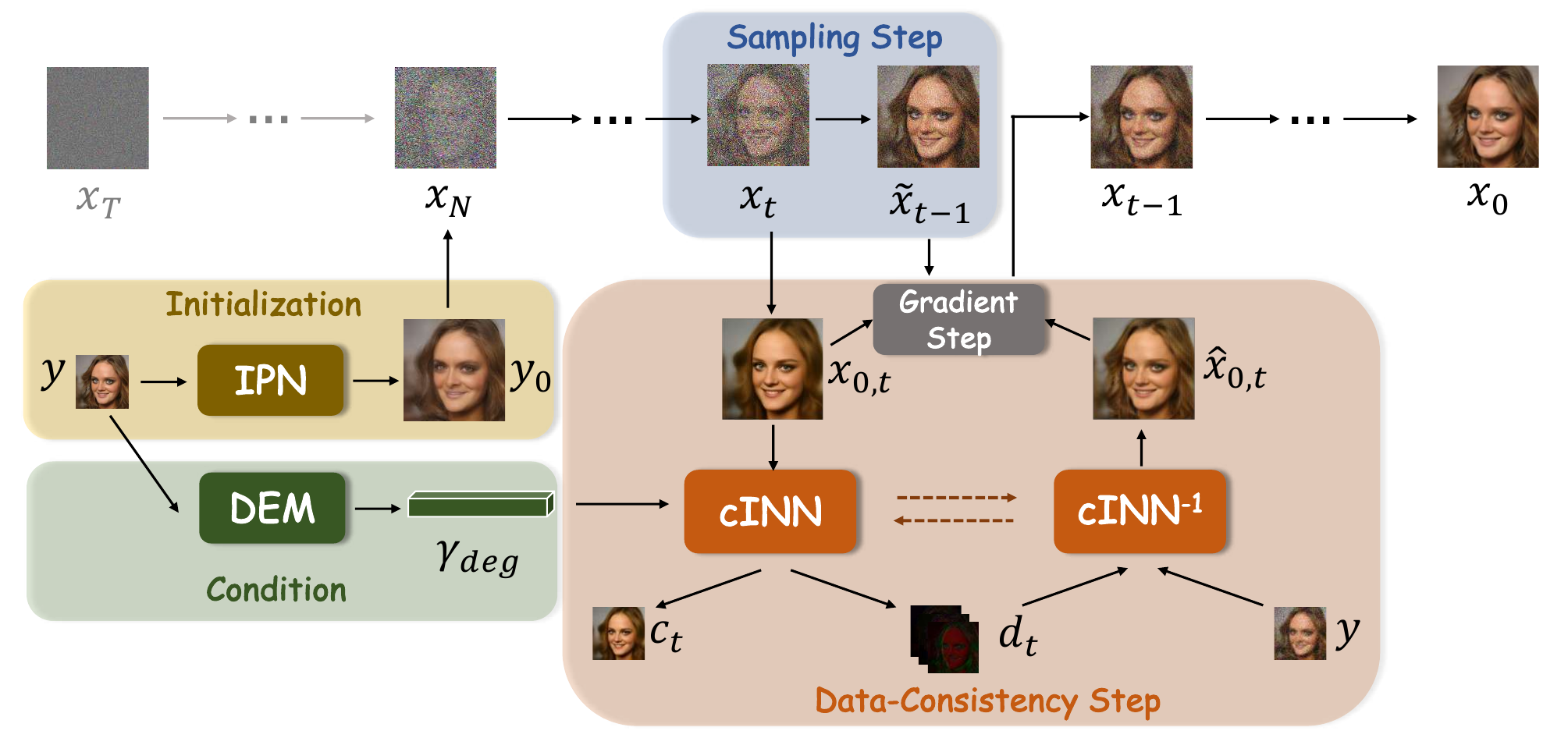}
\caption{ Overview of our BlindINDIGO for blind image restoration. Given a degraded image $\bm{y}$ during inference, our approach first predicts a clean version $\bm{y}_{0}$ with the Initialization Prediction Network (IPN) and extract an implicit degradation embedding $\gamma_{deg}$ with the Degradation Estimation Module (DEM). Next, starting from a diffused $\bm{y}_{0}$, the diffusion posterior sampling is guided by our data-consistency step with INN at each step $t$. We show the detailed algorithm in Algorithm~\ref{algo:blind}.  }
    \label{fig:blind}
\end{figure*}

For a general image restoration problem $\bm{y} = \mathcal{H}(\bm{x},\bm{n})$,
we aim to obtain an image $\tilde{\bm{x}}$ that ensures data consistency while maintaining realistic textures. 
To simultaneously achieve these two goals, we leverage the merit of the perfect reconstruction property of INN and the strong generative prior of pretrained diffusion models. An overview of the proposed approach is shown in Fig.~\ref{fig:non-blind} and Fig.~\ref{fig:blind}. We first train our INN so that its forward part $[\bm{c}, \bm{d}]=f_{\phi}(\bm{x})$ decomposes an image $\bm{x}$ in a coarse and detail part so that 
$\bm{c} \approx \mathcal{H}(\bm{x},\bm{n})$. In other words, $f_{\phi}(\cdot)$ is trained to mimic the degradation process $\mathcal{H}$. Then during the diffusion posterior sampling process, we impose an additional data consistency step after each original unconditional sampling update. 
Specifically, we first utilize our pretrained INN to decompose the intermediate result $\bm{x}_{0,t}$ into the coarse part $\bm{c}_{t}$ that should approximate the degraded measurements and the detail part $\mathbf{d}_{t}$ that models the details lost
during the degradation. We then replace $\bm{c}_{t}$ with the given observed measurements $\bm{y}$. 
Next, the INN-optimized image ${\hat{\bm{x}}}_{0,t}$ is constructed by inverse transform $f_{\phi}^{-1}(\cdot)$ of INN. Therefore, this INN-optimized result ${\hat{\bm{x}}}_{0,t}$ guides the sampling towards satisfying the consistency constraint. 
Simultaneously, ${\hat{\bm{x}}}_{0,t}$ maintains rich details obtained by diffusion posterior sampling without affecting data consistency. Then, the diffusion posterior sampling at the following step is guided by our data-consistent result, ${\hat{\bm{x}}}_{0,t}$, through a gradient operation. Due to the fact that we train an INN to model the degradation process, our algorithm is more flexible than other methods and also more effective given that the invertibility property of the INN ensures that we compute implicitly the equivalent of an inverse at each iteration. 

In the following subsections, we will explain in details how our approach can solve non-blind and blind inverse problems, respectively.

\begin{figure}[t]
    \centering
    \includegraphics[width=0.48\textwidth]{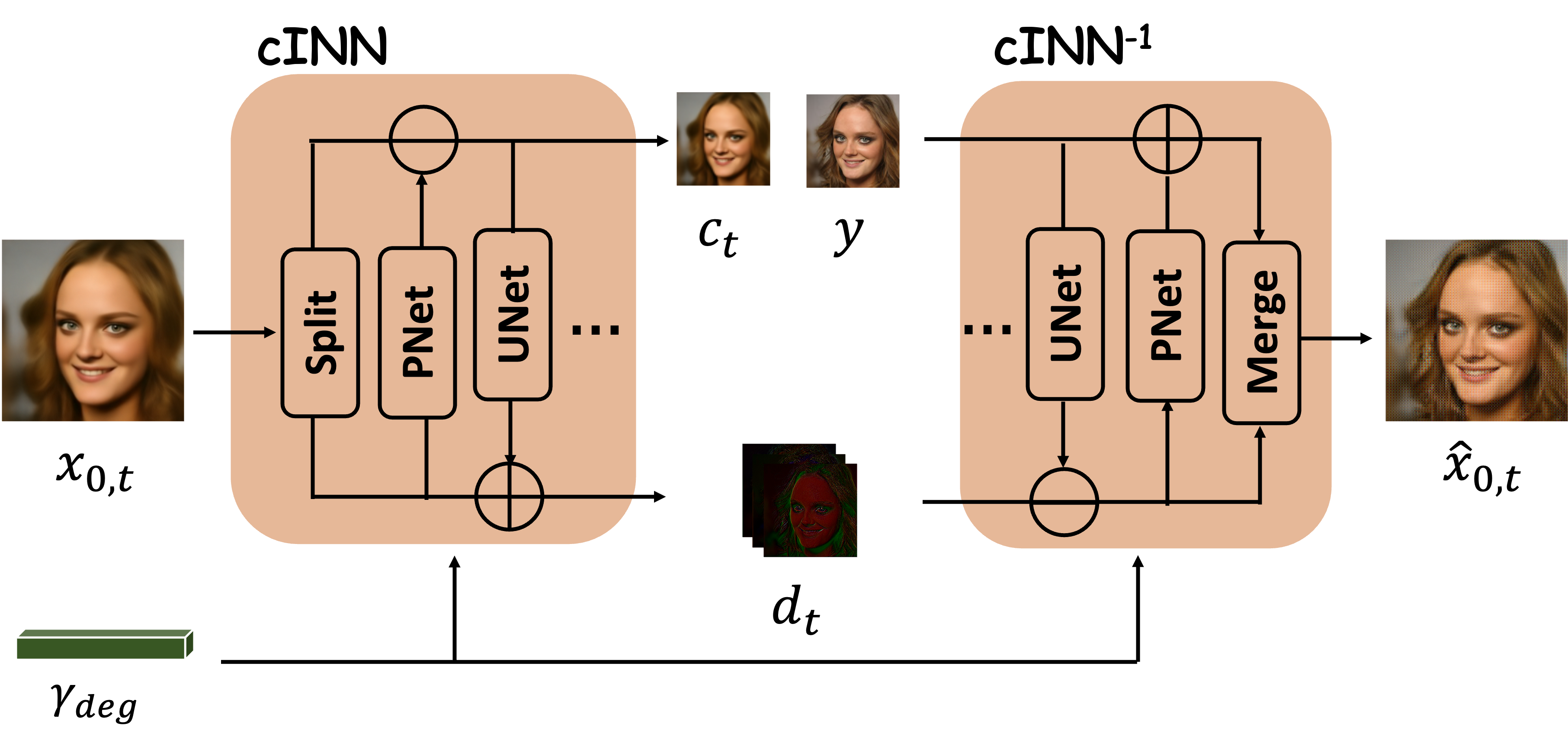}
\caption{ The forward and inverse transform of our INN during inference.}
    \label{fig:INN_framework}
\end{figure}

\begin{figure}[t]
    \centering
    \includegraphics[width=0.48\textwidth]{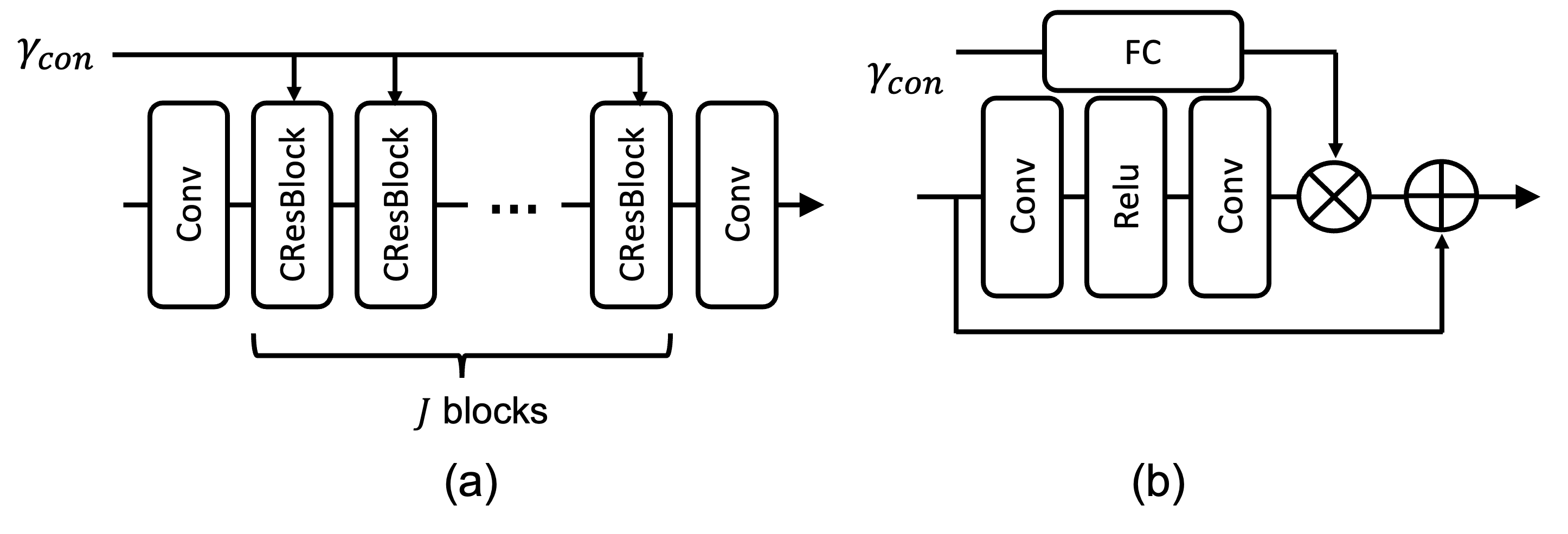}
\caption{(a) The architecture of the PNet/UNet. (b)The architecture of the CResBlcok \cite{cresmd}.  }
    \label{fig:details}
\end{figure}

\subsection{INDIGO for Non-Blind Image Restoration}
\label{non-blind indigo}
In this subsection, we start with non-blind inverse problems and introduce the design of our INN and how it works in the diffusion process.

\textbf{Modelling the degradation process with INN:} 
By exploiting the invertibility of INN, we propose to treat its forward transform $f_{\phi}$ as a simulator of the degradation process and treat its inverse transform $f_{\phi}^{-1}$ as the reconstruction process. To realize this framework, we start with adopting the lifting-inspired invertible blocks in \cite{winnet} (as in Section \ref{background_inn}), which can be expressed as follows:
\begin{align} 
 [\bm{c},\bm{d}] =f_{\phi}(\bm{x}), \qquad
\bm{x}=f_{\phi}^{-1}(\bm{c},\bm{d}),
    \label{eq:LINNbackward}
\end{align}
where the forward transform of INN generates the coarse and detail parts, $\bm{c}$ and $\bm{d}$, while the inverse transform of INN can perfectly recover the input original image from $\bm{c}$ and $\bm{d}$. To model the degradation process,  we impose that $\bm{c}$ resembles $\bm{y}$. 
Given a training set $\left \{ \bm{x}^{i}, \bm{y}^{i}\right \}_{i=1}^{N}$, which contains $N$ high-quality images and their low-quality counterparts, we optimize our INN with the following loss function: 
\begin{align}
\begin{split}
L\left (\phi  \right )=\frac{1}{N}\sum_{i=1}^{N}\left \| {\color{black}{f_{\phi}^{c}(\bm{x}^{ i })}}-\bm{y}^{ i } \right \|_{2}^{2},
\end{split}
\end{align} 
where $\phi$ denotes the set of learnable parameters of our INN and {\color{black}{$f_{\phi}^{c}(\bm{x}^{ i })$ and $f_{\phi}^{d}(\bm{x}^{ i })$ denote the first and second part of the output of $f_{\phi}(\bm{x}^{ i })$, respectively.}} Once we constrain one part of the output of $f_{\phi}(\bm{x}^{ i })$ to be close to $\bm{y}$, due to invertibility, the other part of the output will inevitably represent the detailed information lost during the degradation process.

\textbf{Sampling with the guidance of pretrained INN:}
In the unconditionally trained DDPM \cite{ddpm}, the reverse diffusion process iteratively samples $\bm{x}_{t-1}$ from $p(\bm{x}_{t-1}|\bm{x}_{t})$ to yield clean images $\bm{x}_{0} \sim q(\bm{x})$ from initial random noise $\bm{x}_{T} \sim \mathcal{N}(\mathbf{0},\mathbf{I})$. 
{\color{black}{Here, we rewrite Eq.~\ref{eq:ddpm reverse 1} with the pre-trained approximator $\boldsymbol{\epsilon}_{\boldsymbol{\theta}}(\bm{x}_{t},t)$ and split it into the following two equations:
\begin{align}
\begin{split}
{\color{black}{\bm{x}_{0,t}  = \frac{1}{\sqrt{\bar\alpha_t}}(\bm{x}_{t} - \sqrt{1 - \bar\alpha_t} \bepsilon_\theta(\bm{x}_t, t) )}}
\end{split}
\label{X0}
\end{align}
and 
\begin{align}
\begin{split}
{\color{black}\bm{x}_{t-1}=
{\frac{\sqrt{\alpha_t}(1-\bar\alpha_{t-1})}{1 - \bar\alpha_t}\bm{x}_{t}+\frac{\sqrt{\bar\alpha_{t-1}}\beta_t}{1 - \bar\alpha_t}\bm{x}_{0,t}  + \sigma_t \bz}}.
\end{split}
\label{xt-langevin}
\end{align}
As illustrated in Eq. \ref{X0}, $\bm{x}_{0,t}$ is the predicted clean image from the noisy image $\bm{x}_{t}$.}} {\color{black}{To solve inverse problems, we need to refine each unconditional transition using $\bm{y}$ to ensure data consistency. }}In our proposed algorithm, we impose our data-consistency step by modifying the clean image $\bm{x}_{0,t}$ instead of the noisy image $\bm{x}_{t}$.

As shown in Algorithm~\ref{algo:nonblind}, we impose an additional data consistency step (in blue) with our off-the-shelf INN after each original unconditional sampling update. 
In this additional step, we apply the forward transform $f_{\phi}(\cdot)$ to the intermediate result $\bm{x}_{0,t}$ leading to the decomposition of $\bm{x}_{0,t}$ into coarse and detail part $\bm{c}_{t}$, $\bm{d}_{t}$ respectively. We then replace the coarse part $\mathbf{c}_{t}$ with the measurements $\bm{y}$. The INN-optimized $\hat{\bm{x}}_{0,t}$ is then generated by applying the inverse transform $f_{\phi}^{-1}(\cdot)$ to $\{\bm{y},\bm{d}_{t}\}$. Thus, the INN-optimized $\hat{\bm{x}}_{0,t}$ is composed of the coarse information $\bm{y}$ and the details generated by the diffusion process. To incorporate the INN-optimized $\hat{\bm{x}}_{0,t}$ into the DDPM algorithm, 
we update $\bm{x}_{t}$ with the guidance of the gradient of $\|{\hat{\bm{x}}_{0,t} - \bm{x}_{0,t}}\|_2^2$. With the help of INN, our algorithm effectively estimates the details lost in the degradation process and is no longer limited by the requirement of knowing the exact expression of the degradation model, since the degradation is learned through data.

\begin{algorithm}[t]
\small
	\caption{{Non-Blind INDIGO}}

	\KwIn{Corrupted image $\boldsymbol{y}$, gradient scale $\zeta$, INN $f_{\phi}$.}
	\KwOut{Output image $\boldsymbol{x}_{0}$ conditioned on $\boldsymbol{y}$}
    $\bm{x}_T \sim \mathcal{N}(\bzero, \bI)$
    
	\For{$t$ from $T$ to 1}{
        $\bz \sim \mathcal{N}(\bzero, \bI)$ if $t > 1$, else $\bz = \bzero$
        
        $\bm{x}_{0,t}  = \frac{1}{\sqrt{\bar\alpha_t}}(\bm{x}_{t} - \sqrt{1 - \bar\alpha_t} \bepsilon_\theta(\bm{x}_t, t) )$
        
        $\tilde{\bm{x}}_{t-1}  = \frac{\sqrt{\alpha_t}(1-\bar\alpha_{t-1})}{1 - \bar\alpha_t}\bm{x}_{t}+\frac{\sqrt{\bar\alpha_{t-1}}\beta_t}{1 - \bar\alpha_t}\bm{x}_{0,t}  + \sigma_t \bz$
        
        {\color{blue}{$\bm{c}_{t},\bm{d}_{t} =f_{\phi}(\bm{x}_{0,t})$}}
        
        {\color{blue}{$\hat{\bm{x}}_{0,t} =f_{\phi}^{-1}(\bm{y},\bm{d}_{t})$}}

        {\color{blue}{$\bm{x}_{t-1} =\tilde{\bm{x}}_{t-1}  - { {\zeta}}\nabla_{\bm{x}_{t}} \|{\hat{\bm{x}}_{0,t} - \bm{x}_{0,t}}\|_2^2$}}

	}
	\Return $\boldsymbol{x}_{0}$
\label{algo:nonblind}
\vspace{-0.1em}
\end{algorithm}

\subsection{BlindINDIGO for Blind Image Restoration}
\label{blind}
{In the previous section, our INN is learned using a fully supervised approach given that we assume to have access to a training set $\left \{ \bm{x}^{i}, \bm{y}^{i}\right \}_{i=1}^{N}$ where the degradation in $ \bm{y}^{i}$ is fully consistent with the degradation of the actual measurements.} In practical scenarios, many images undergo complex and unknown degradation processes. Some works \cite{blinddps,gibbsddrm} solve this by assuming a closed-form expression of the degradation process and then they predict the parameters in this expression. In this work, by simulating several degradation processes and through finetuning, our approach can deal with unknown, linear and non-linear degradation processes. The algorithm is described in Fig.~\ref{fig:blind} and Algorithm~\ref{algo:blind}.

\textbf{Conditonal INN for blind image restoration:} 
Since one set of parameters $\phi$ in $f_{\phi}(\bm{x})$ can simulate one type of degradation, we take different degradation labels as an additional input to guide the forward and inverse transform of INN, i.e., $f_{\phi}(\bm{x},\bm{\gamma}_{deg})$, {to simulate multiple different degradation processes}.
{To extract the degradation information $\bm{\gamma}_{deg}$, we utilize a pre-trained Degradation Estimation Module (DEM) to model the degradation implicitly in the latent feature space, since real-world degradations are usually too complex to be modeled with an explicit combination of multiple degradation types.}
As depicted in Fig. \ref{fig:INN_framework}, we keep the basic invertible blocks of \cite{winnet} and modulate them with the degradation vector $\bm{\gamma}_{deg}$. Specifically, the image is split into two parts by a splitting operator. Then the Prediction Network (PNet) conditioned on the coarse part aims to predict the detail part, while the Update Network (UNet) conditioned on the detail part is used to adjust the coarse part to make it smoother. The Prediction and Update networks are applied alternatively to generate the coarse and detail parts $\bm{c}$ and $\bm{d}$. The details of how the degradation vector $\bm{\gamma}_{deg}$ controls the features are shown in Fig. \ref{fig:details}.  The degradation vector $\bm{\gamma}_{deg}$ is passed through a fully-connected layer and a channel-wise multiplication is added to the original Residual block \cite{resnet} to control the summation weight. To train this conditional INN, the loss function becomes:
\begin{align}
\begin{split}
\label{eq:lossfunc}
L\left ( \phi  \right )=\frac{1}{N}\sum_{i=1}^{N}\left \| {\color{black}{f_{\phi}^{ c }}}(\bm{x}^{ i },\bm{\gamma}_{deg}^{ i } )-\bm{y}^{ i } \right \|_{2}^{2},
\end{split}
\end{align} 
{\color{black}{where $f_{\phi}^{c}(\bm{x}^{ i },\bm{\gamma}_{deg}^{ i } )$ denotes the first part of the output of $f_{\phi}(\bm{x}^{ i },\bm{\gamma}_{deg}^{ i } )$ {and the implicit degradation vector $\bm{\gamma}_{deg}^{ i }$
is generated by a pretrained DEM $h_{\kappa}(\cdot)$.}}} Here, the training set $\left \{ \bm{x}^{i}, \bm{y}^{i}, \bm{\gamma}_{deg}^{ i }\right \}_{i=1}^{N}$ contains $N$ high-quality images,  their low-quality counterparts, and the implicit degradation vector $\bm{\gamma}_{deg}^{ i }$ = $h_{\kappa}(\bm{y}^{ i })$ generated by a pretrained DEM.
\begin{algorithm}[t]
\small
	\caption{{BlindINDIGO}}

	\KwIn{Corrupted image $\boldsymbol{y}$, gradient scale $\zeta$, pretrained INN $f_{\phi}(\cdot)$, pretrained IPN $g_{\omega}(\cdot)$ , implicit degradation embedding $\bm{\gamma}_{deg}$ extracted by pre-trained DEM from $\boldsymbol{y}$, learning rate $l$ for optimizing INN.}
	\KwOut{Output image $\boldsymbol{x}_{0}$ conditioned on $\boldsymbol{y}$}

         {\color{blue}{$\boldsymbol{\eta} \sim \mathcal{N}(\bzero, \bI)$}}

          {\color{blue}{$\bm{x}_N = \sqrt{\Bar{\alpha}_{N}}g_{\omega}(\bm{y})+ \sqrt{1-\Bar{\alpha}_{N}} \boldsymbol{\eta} $}}

	\For{$t$ from $N$ to 1}{
        $\bz \sim \mathcal{N}(\bzero, \bI)$ if $t > 1$, else $\bz = \bzero$
        
        $\bm{x}_{0,t}  = \frac{1}{\sqrt{\bar\alpha_t}}(\bm{x}_{t} - \sqrt{1 - \bar\alpha_t} \bepsilon_\theta(\bm{x}_t, t) )$
        
        $\tilde{\bm{x}}_{t-1}  = \frac{\sqrt{\alpha_t}(1-\bar\alpha_{t-1})}{1 - \bar\alpha_t}\bm{x}_{t}+\frac{\sqrt{\bar\alpha_{t-1}}\beta_t}{1 - \bar\alpha_t}\bm{x}_{0,t}  + \sigma_t \bz$

        {\color{blue}{$\bm{c}_{t},\bm{d}_{t} =f_{\phi}(\bm{x}_{0,t}, \bm{\gamma}_{deg})$}}
        
        {\color{blue}{$\hat{\bm{x}}_{0,t} =f_{\phi}^{-1}(\bm{y},\bm{d}_{t},  \bm{\gamma}_{deg})$}}

        {\color{blue}{$L_{total}=\lambda_{F}L_{F}({\bm{c}_{t}, \bm{y}})+\lambda_{I}L_{I}({\hat{\bm{x}}_{0,t}, \bm{x}_{0,t}})$}}

         {\color{blue}{$\bm{x}_{t-1} =\tilde{\bm{x}}_{t-1}  - { {\zeta}}\nabla_{\bm{x}_{t}} L_{total}$}}

        {\color{blue}{$\phi \leftarrow \phi - l \nabla_{\phi} \|{\bm{c}_{t}- \bm{y}}\|_2^2$}}

	}
	\Return $\boldsymbol{x}_{0}$
\label{algo:blind}
\vspace{-0.1em}
\end{algorithm}

\renewcommand{\arraystretch}{1.2} 
\begin{table*}[]
    \caption{Quantitative (PSNR$\uparrow$/FID$\downarrow$/LPIPS$\downarrow$/NIQE$\downarrow$) comparison on 4${\times}$ SR with different levels of Gaussian noise.
    \textbf{Bold} texts represent the best performance. }
    \label{tab:nonblind}
    \centering
        \begin{tabular}{|l| |c|c|c|c||c|c|c|c||c|c|c|c|}
\hline 
            \multirow{2}{*}{Methods }                              & \multicolumn{4}{c||}{Noise $\sigma$=0}& \multicolumn{4}{c||}{Noise $\sigma$=0.05} &  \multicolumn{4}{c|}{Noise $\sigma$=0.10} \\ \cline{2-13}                             & PSNR$\uparrow$ & FID$\downarrow$ &LPIPS$\downarrow$  & NIQE$\downarrow$ &  PSNR$\uparrow$ & FID$\downarrow$ &LPIPS$\downarrow$  & NIQE$\downarrow$    & PSNR$\uparrow$ & FID$\downarrow$ &LPIPS$\downarrow$  & NIQE$\downarrow$  \\ \hline \hline 
ILVR~\cite{ilvr}                      & 27.43&44.04 &0.2123 &5.4689 &26.42 & 60.27&0.3045 & 4.6527 &24.60 &88.88 &0.4833 &4.4888       \\  
 DDRM~\cite{DDRM}                        &28.08&65.80 &0.1722& 4.4694&{27.06} &45.90  &0.2028& 4.8238&26.16 & 45.49 & 0.2273 & 4.9644   \\  
 DPS~\cite{dps}                    & 26.67 &32.44 & 0.1370 &4.4890 &25.92&31.71 &0.1475&4.3743 &  24.73 &31.66 &0.1698&4.2388\\  
  \hline  
 \textbf{Ours}                 & \textbf{28.15}     &\textbf{22.33} & \textbf{0.0889 }& \textbf{4.1564}&{27.16}& \textbf{26.64}&\textbf{0.1215}&\textbf{4.1004}&\textbf{26.25} &\textbf{28.89} &\textbf{0.1399}& \textbf{3.9659} \\   
  {\textbf{Ours-DDIM}        }         & {28.06}     &{24.51} & {0.0966}& {4.2524}&{\textbf{27.19}}& {28.07}&{0.1271}&{4.3619} &{26.22} &{30.85}&{0.1488} &{4.2189} \\  
\hline  

    \end{tabular}
\end{table*}
\begin{figure}[t]
    \centering
    \includegraphics[width=0.48\textwidth]{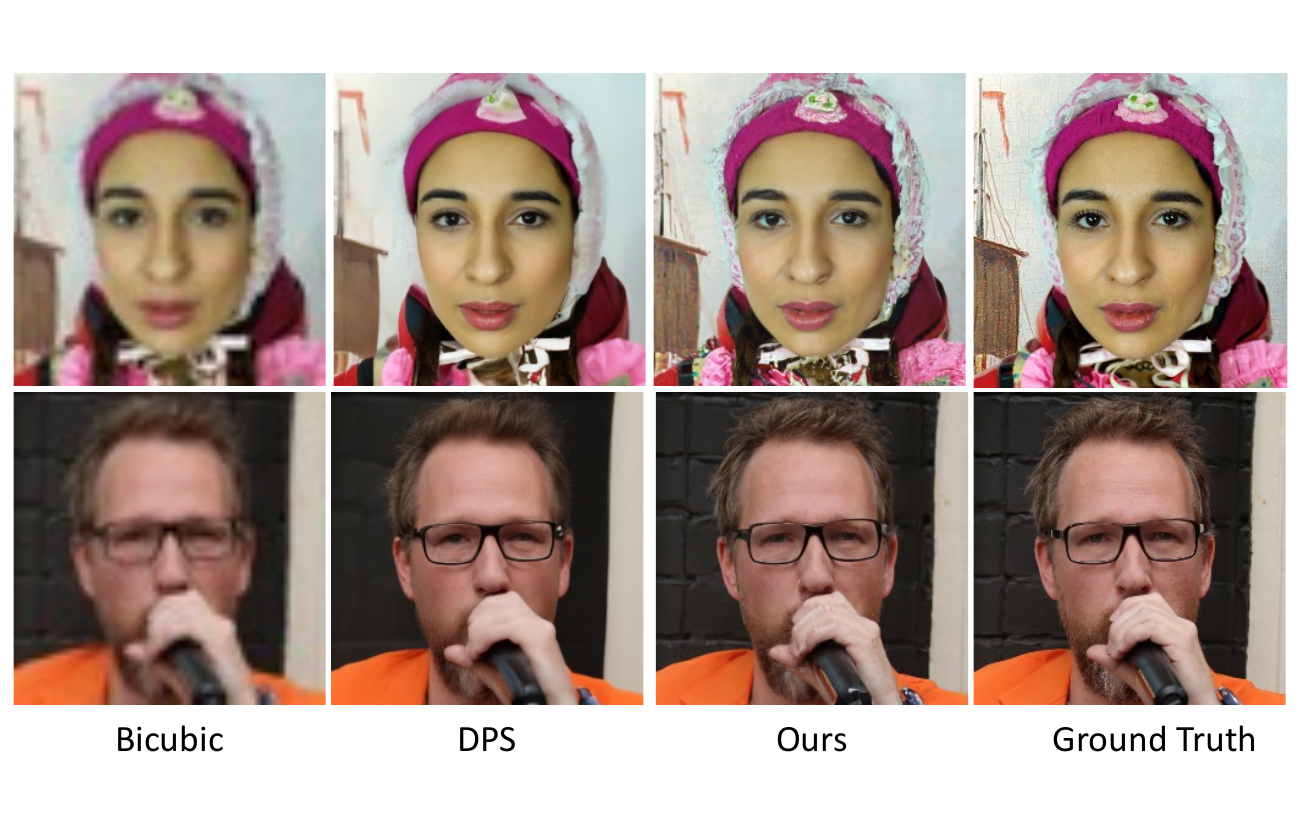}
        \vspace{-0.5cm}
\caption{
Comparisons with state-of-the-art image restoration approach \cite{dps} on solving the \textbf{non-blind} super-resolution problem (x4)  on FFHQ validation dataset.}
    \label{fig:com}
\end{figure}

\begin{figure}[t]
    \centering
    \includegraphics[width=0.5\textwidth]{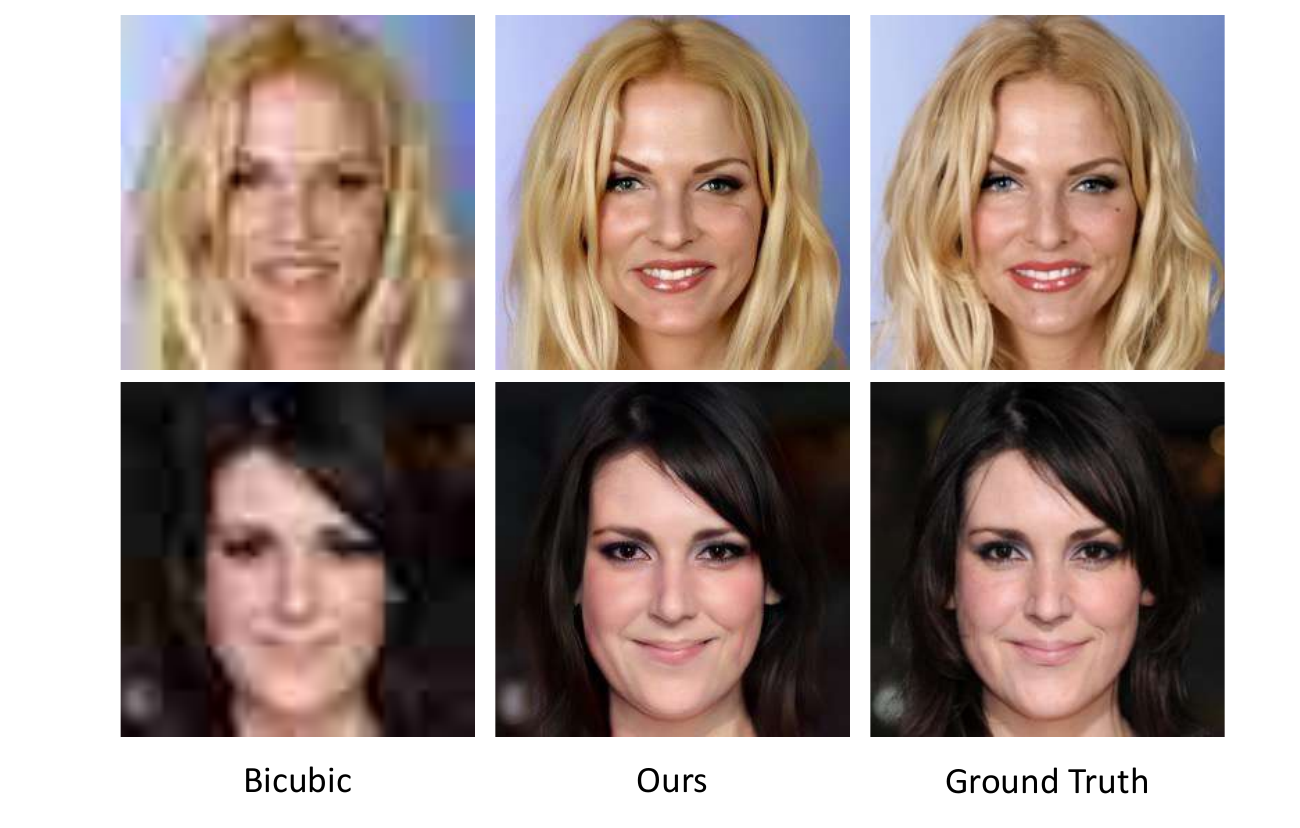}
\caption{Results of our algorithm on solving the \textbf{non-blind} inverse problem with Jpeg compression on CelabA HQ validation dataset.}
    \label{fig:jpeg10}
     \vspace{-0.4cm}
\end{figure}
\begin{figure}[t]
    \centering
    \includegraphics[width=0.45\textwidth]{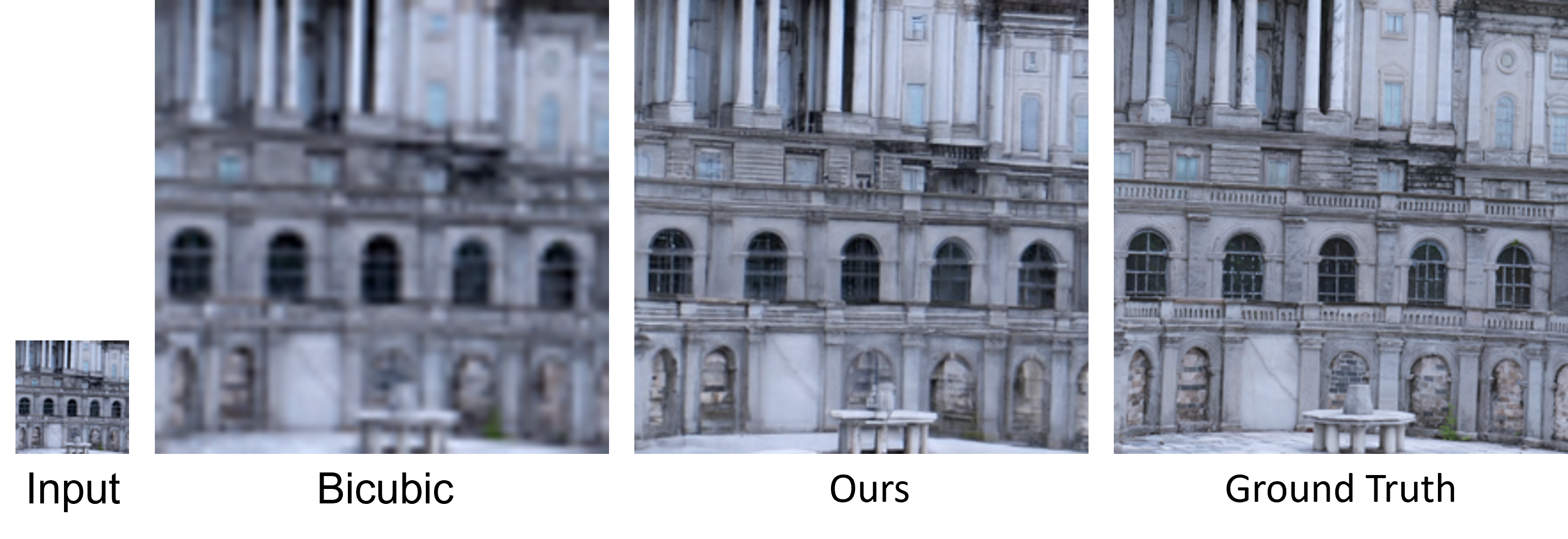}
\caption{ {\color{black}{Result of our algorithm on reconstructing real images from DRealSR \cite{drealsr} with resolution enhancement by a factor 4 per direction (\textbf{non-blind}). 
}}}
    \label{fig:real}
\end{figure}
\textbf{Guiding posterior sampling with INN:} 
Similar to non-blind INDIGO, we still apply the forward transform $f_{\phi}(\cdot)$ to the intermediate result $\bm{x}_{0,t}$ and then replace its coarse part $\mathbf{c}_{t}$ with the measurements $\bm{y}$. The INN-optimized $\hat{\bm{x}}_{0,t}$ is then generated by applying the inverse transform $f_{\phi}^{-1}(\cdot)$.
The invertibility of the INN allows us to compute the gradient step in either the measurement or the image domain. In the non-blind case, we only operate in the image domain. 
For the blind case,
to further improve the reconstruction performance, we operate in both domains. Specifically, we take the gradient of the following loss:
\begin{align}
\begin{split}
\label{eq:lossfunc}
L_{total}=\lambda_{F}L_{F}({\bm{c}_{t}, \bm{y}})+\lambda_{I}L_{I}({\hat{\bm{x}}_{0,t}, \bm{x}_{0,t}}),
\end{split}
\end{align} 
where we constrain the measurement space  with $L_{F}({\bm{c}_{t}, \bm{y}})$ and the high-quality image space with $L_{I}({\hat{\bm{x}}_{0,t}, \bm{x}_{0,t}})$. Here, $\lambda_{F}$ and $\lambda_{I}$ denote the loss weights of $L_{F}$ and $L_{I}$, respectively. In our implementation, we set $\lambda_{F}=2.5, \lambda_{I}= 1$, {\color{black}{$ L_{F}({\bm{c}_{t}, \bm{y}})=\|\bm{c}_{t}-\bm{y}\|_{2}^{2}$ and $  L_{I}({\hat{\bm{x}}_{0,t}, \bm{x}_{0,t}})=\|V(\hat{\bm{x}}_{0,t})-V(\bm{x}_{0,t})\|_{2}^{2}$, where $V(\cdot)$ denotes the feature embedding space of VGG16 \cite{vgg} network.}}  We discuss the detailed implementation and ablation study in Section \ref{loss_ablation}.

\textbf{Finetuning our INN during sampling:} By replacing the INN in Algorithm~\ref{algo:nonblind} with the above pretrained conditional INN, our approach can deal with multiple inverse problems. However, both conditional INN and the DEM are trained with synthetic degradation data pairs that may not model exactly the actual degradation. In real-world scenarios with more complex degradations, the parameters of our INN need to be refined to simulate the degradation processes more accurately. We achieve this by finetuning the parameters in INN at testing stage. This is done at the end of each iteration as shown in Algorithm~\ref{algo:blind}. 
{In this step given the current estimated image $\bm{x}_{0,t}$, the parameters of our INN, $f_{\phi}$, are updated to reduce $L_{F}({\bm{c}_{t}, \bm{y}})$ in  Eq.~(\ref{eq:lossfunc}).}

{\textbf{Accelerating our algorithm with initialization:} To accelerate our algorithm by reducing the number of timesteps, we introduce an initialization strategy. As observed in \cite{ccdf,yue2022difface}, starting from a single forward diffusion with better initialization instead of Gaussian noise significantly reduces the number of sampling steps in the diffusion posterior sampling process. Following \cite{yue2022difface}, 
and as shown
in Algorithm~\ref{algo:blind}, we first produce an initial restored image, $g_{\omega}(\bm{y})$, using an image restoration method. By construction, this image
would already look consistent with the measurements but lacks details. Then it is forward-diffused (noise-added) to generate the starting point of sampling, $\bm{x}_N = \sqrt{\Bar{\alpha}_{N}}g_{\omega}(\bm{y})+ \sqrt{1-\Bar{\alpha}_{N}} \boldsymbol{\eta}$. In our setting, $g_{\omega}(\cdot)$ is the SwinIR \cite{liang2021swinir} method trained with L2 Loss.}

{\subsection{Algorithm~\ref{algo:nonblind} vs. Algorithm~\ref{algo:blind}:} In summary, Algorithm~\ref{algo:nonblind} can be used in fully supervised settings and for the case of 
the single, fixed degradation. It requires training data pairs which are degraded in the same way as the testing dataset to train the INN. To evaluate Algorithm~\ref{algo:nonblind}, we need to train different INNs for different degradations. Algorithm~\ref{algo:blind} can be used for multiple degradations including unseen cases. For example, to reconstruct measurement $\bm{y}$ from degradation model $\mathcal{H}$ during inference, in the non-blind case (Algorithm~\ref{algo:nonblind}) we have access to labeled dataset $\left \{ \bm{x}^{i}, \bm{y}^{i}=\mathcal{H}( \bm{x}^{i})\right \}_{i=1}^{N}$ to train our INN, while in the blind case (Algorithm~\ref{algo:blind}) we need to generate a dataset with several distortion models: $\left \{ \bm{x}^{i}, \bm{y}^{i}=\mathcal{H}^{j}( \bm{x}^{i})\right \}_{i=1}^{N}$, 
$\mathcal{H}^{j}\in \{\mathcal{H}^{j}\}_{j=1}^{J}$
to train our INN and the correct $\mathcal{H}$ may not belong to $\{\mathcal{H}^{j}\}_{j=1}^{J}$.}

\section{Experimental Results}
\subsection{Results on Non-blind Image Restoration}
\subsubsection{Implementation Details}\label{exp_nonblind_implement}
{\color{black}{Empirically, we set $\zeta$=0.5 and $T$=1000 in Algorithm~\ref{algo:nonblind}. To implement our INN, we follow the structure of the invertible blocks in \cite{winnet}. Specifically, our INN consists of 2 levels of the lifting-inspired invertible blocks and each block is constructed using the same set of 4 pairs of PNet and Unet as in \cite{winnet}. Each PNet/Unet network consists of an input convolutional layer, 2 residual blocks with depth-wise separable convolution layers, and an output convolutional layer. The number of feature channels in PUNet is set to 32 and the spatial filter size in depth-wise separable convolutional layers is set to 5. }}The total number of learnable parameters of our INN is 0.71M. 

We test our method on the FFHQ 256×256 1k validation dataset \cite{ffhq}, CelebA HQ 256×256 1k validation dataset \cite{celeba}, and a real-world SR dataset DRealSR \cite{drealsr}. For the face photo reconstruction, we utilize a pre-trained unconditional diffusion model trained on the FFHQ training dataset by \cite{dps} and select 10k images from the FFHQ training dataset to train our INN.
For the natural image reconstruction on \cite{drealsr}, we utilize the pre-trained unconditional diffusion model trained on ImageNet \cite{Imagenet} by \cite{dhariwal2021diffusion} and use DRealSR \cite{drealsr} training dataset to train our INN. 
We apply our proposed method to three settings for inverse problems:  bicubic downsampling with/without noise, non-linear degradation model based on combining downsampling with jpeg compression, real-world degradation model. {\color{black}{In this section, we assume the degradation model is known, so for the first two settings, we synthesize training and testing data using the same degradation. For the third setting, we train our INN with DRealSR training dataset \cite{drealsr}, which is a large-scale diverse SR benchmark obtained by zooming digital single-lens reflex (DSLR) cameras to collect real low-resolution (LR) and high-resolution (HR) images. After simulating its degradation with our INN, we test our algorithm with DRealSR test dataset \cite{drealsr}}}. The reconstruction results are evaluated with PSNR, FID \cite{fid}, LPIPS \cite{lpips} and NIQE \cite{niqe}.

\renewcommand{\arraystretch}{1.2} 
\begin{table*}[]
    \caption{Quantitative (PSNR$\uparrow$/LPIPS$\downarrow$/IDS$\uparrow$) comparison on 4${\times}$ SR with different levels of degradations. \textbf{Bold} texts represent the best performance. }
    \label{tab:blind}
    \centering
        \begin{tabular}{|l| |c|c|c||c|c|c||c|c|c|}
\hline 
            \multirow{2}{*}{Methods }                              & \multicolumn{3}{c||}{Mild}& \multicolumn{3}{c||}{Medium} &  \multicolumn{3}{c|}{Severe} \\ \cline{2-10}                             & PSNR$\uparrow$ & LPIPS$\downarrow$  & IDS$\uparrow$ &  PSNR$\uparrow$ & LPIPS$\downarrow$  & IDS$\uparrow$    & PSNR $\uparrow$ & LPIPS$\downarrow$  & IDS$\uparrow$  \\ \hline \hline 
PGDiff~\cite{yang2023pgdiff}           &23.38&0.2668&0.9509&   22.56&0.2883&0.9442    & 21.96&0.3062 &0.9371           \\  
 DifFace~\cite{yue2022difface}                       & 24.32&0.2782&0.9441   &23.94&0.2896&0.9362       & 23.45&0.3014&0.9270    \\  
 DR2~\cite{wang2023dr2}                    & \textbf{{{25.86}}}&0.2775  & 0.9312 & 24.23 &0.2916 &0.9105       & 23.34&0.3030  &  0.9169  \\  
  StableSR~\cite{stablesr}                    & {{{25.70}}}&0.2155  & 0.9545 & 24.46 &0.2371 &0.9552       & 23.62&0.2559  &  0.9538  \\  \hline 
 \textbf{Ours}                 & {\color{black}{25.41}}&\textbf{0.2142}&{\textbf{0.9662}}       & {{24.85}}&{\textbf{0.2360}}&{\textbf{0.9585}} 
& {\textbf{24.16}}&{\textbf{0.2534}}&{\textbf{0.9558}}  \\ 
{\textbf{Ours-DDIM}}                 &{{{25.72}}}&{{0.2297}}&{{{0.9634}}}       & {{\textbf{25.09}}}&{{{0.2479}}}&{{{0.9570}}}
& {{\textbf{24.31}}}&{{{0.2550}}}&{{{0.9390}}}  \\ 
\hline

    \end{tabular}
\end{table*}
\begin{figure*}[!tp]\footnotesize 
	\centering
\hspace{-0.2cm}
\begin{tabular}{c@{\extracolsep{0.2em}}c@{\extracolsep{0.2em}}c@{\extracolsep{0.2em}}c@{\extracolsep{0.2em}}c@{\extracolsep{0.2em}}c@{\extracolsep{0.2em}}c@{\extracolsep{0.2em}}c@{\extracolsep{0.2em}}c@{\extracolsep{0.2em}}c}
    &\includegraphics[width=0.13\textwidth]{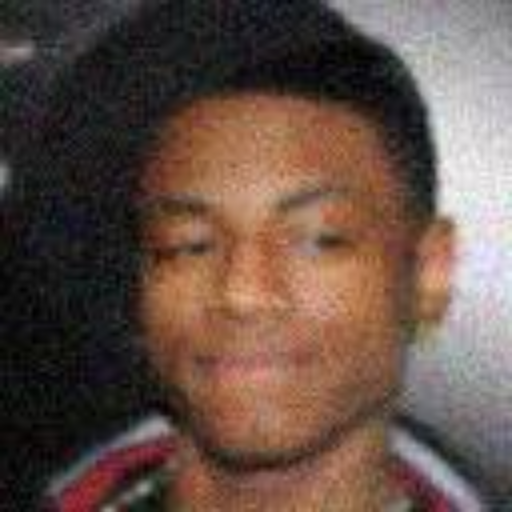}~
		&\includegraphics[width=0.13\textwidth]{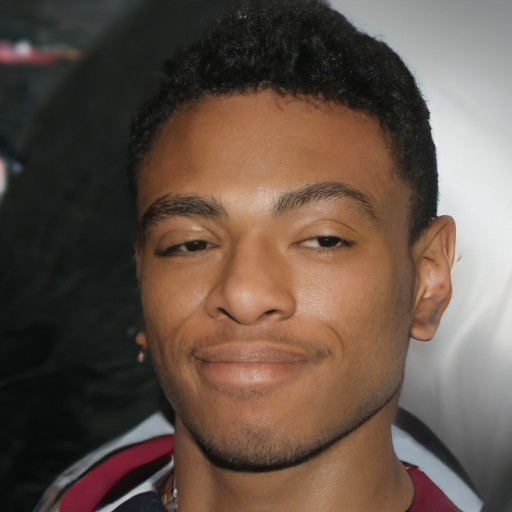}~

    &\includegraphics[width=0.13\textwidth]{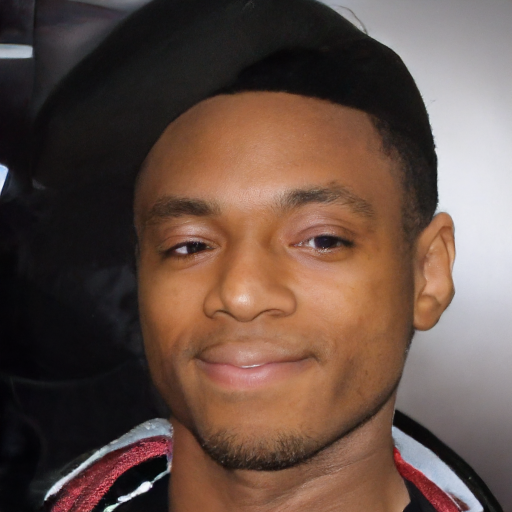}~
        &\includegraphics[width=0.13\textwidth]{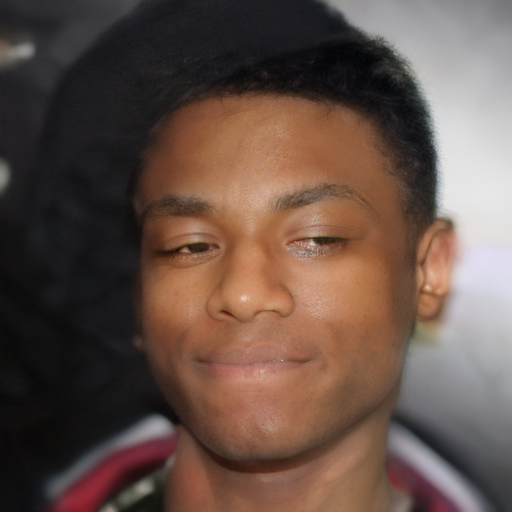}~
        &\includegraphics[width=0.13\textwidth]{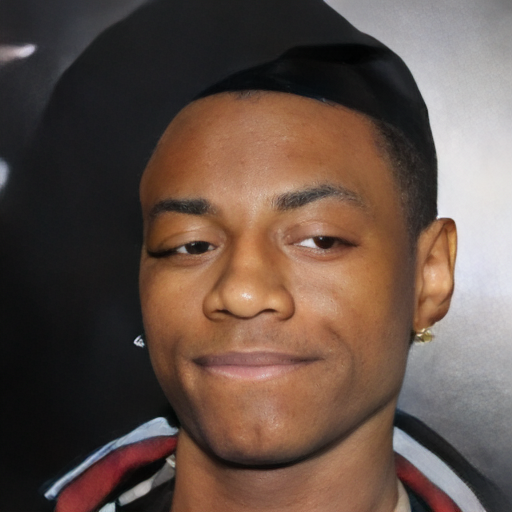}~
		&\includegraphics[width=0.13\textwidth]{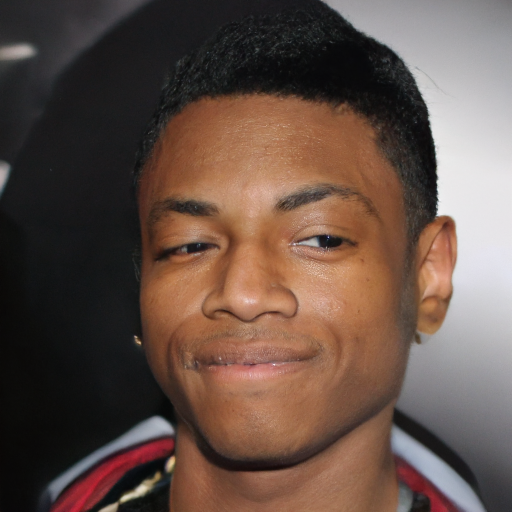}~
		&\includegraphics[width=0.13\textwidth]{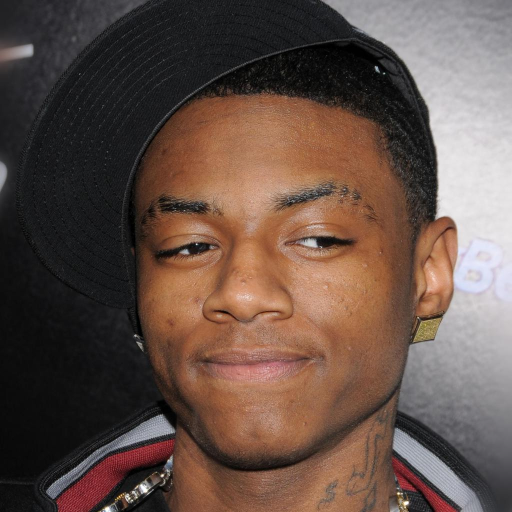}\\
  &\includegraphics[width=0.13\textwidth]{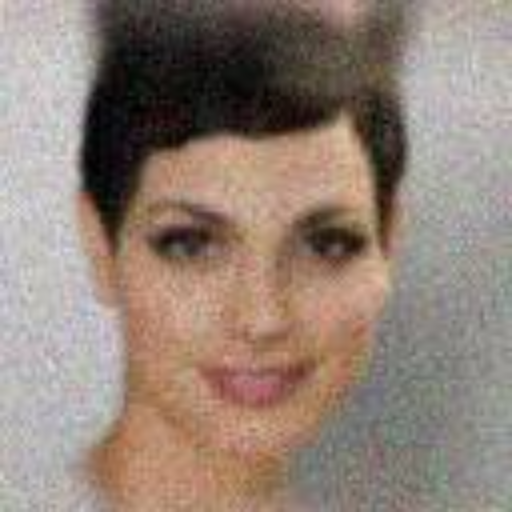}~
		&\includegraphics[width=0.13\textwidth]{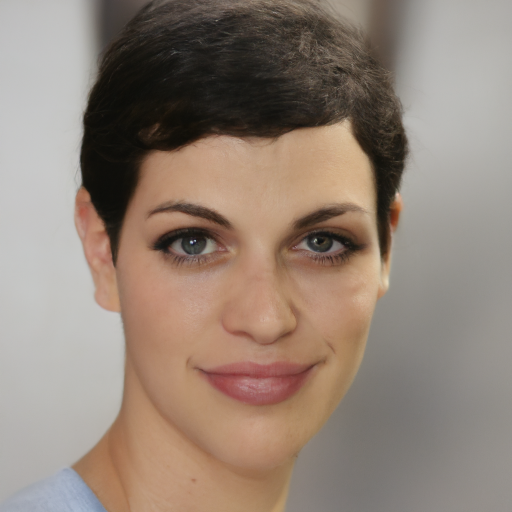}~

    &\includegraphics[width=0.13\textwidth]{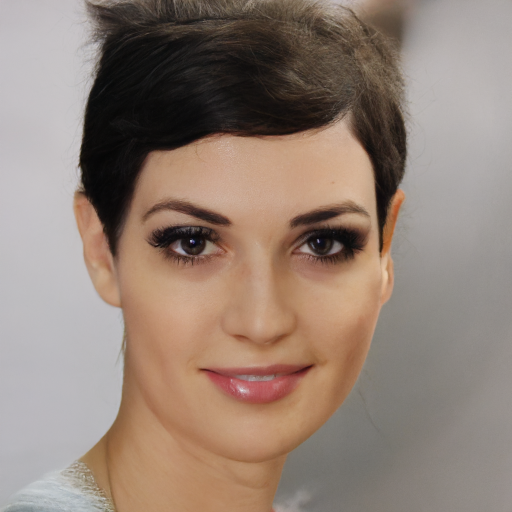}~
        &\includegraphics[width=0.13\textwidth]{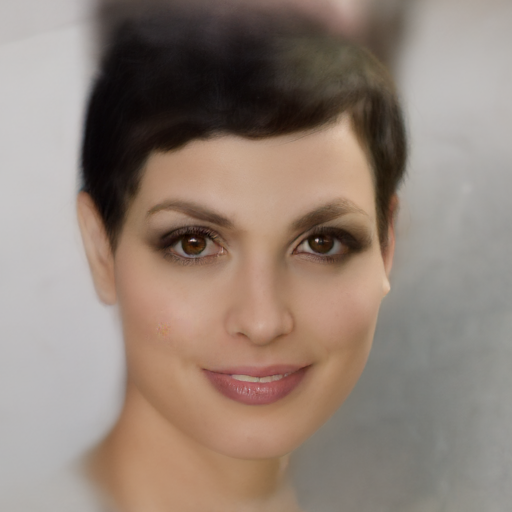}~
        &\includegraphics[width=0.13\textwidth]{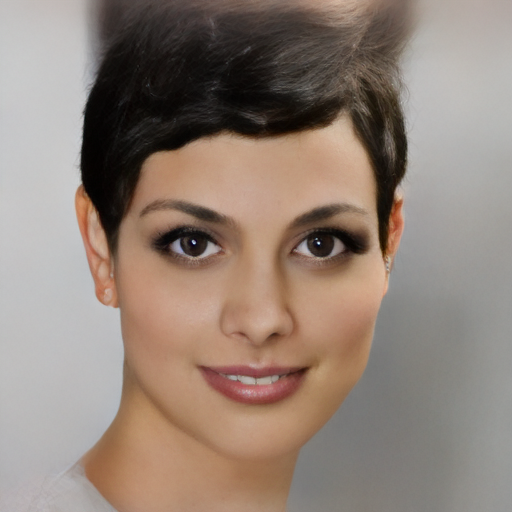}~
		&\includegraphics[width=0.13\textwidth]{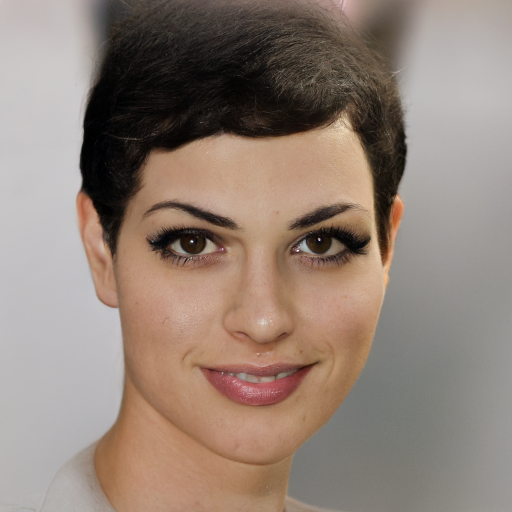}~
		&\includegraphics[width=0.13\textwidth]{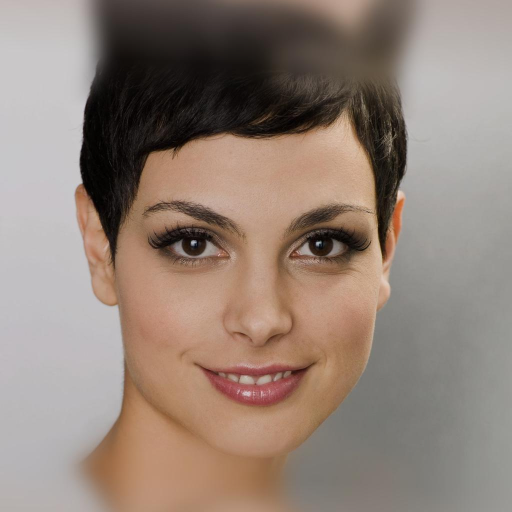}\\
& (a) Input  & (b) DifFace  & (c) PGDiff   & (d) DR2  & (e) {StableSR}& \textbf{(f) Ours} & \textbf{(g) GT}\\
    
	\end{tabular}
	\caption{{Comparisons on 4x \textbf{blind} SR with mild degradation on CelebA-HQ.}} 
	\label{fig:mild}
\end{figure*}
\begin{figure*}[!tp]\footnotesize 
	\centering
\hspace{-0.2cm}
\begin{tabular}{c@{\extracolsep{0.2em}}c@{\extracolsep{0.2em}}c@{\extracolsep{0.2em}}c@{\extracolsep{0.2em}}c@{\extracolsep{0.2em}}c@{\extracolsep{0.2em}}c@{\extracolsep{0.2em}}c@{\extracolsep{0.2em}}c@{\extracolsep{0.2em}}c}
    &\includegraphics[width=0.13\textwidth]{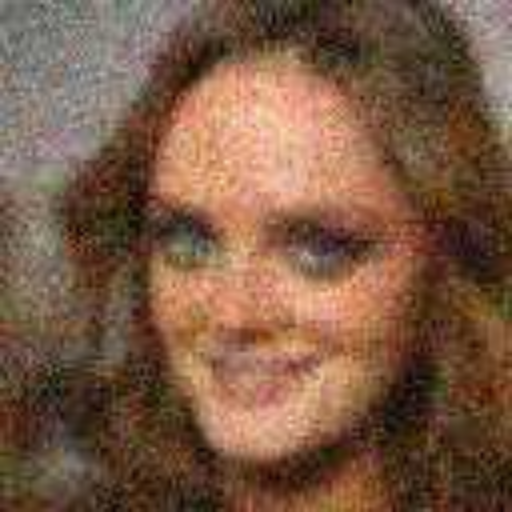}~
		&\includegraphics[width=0.13\textwidth]{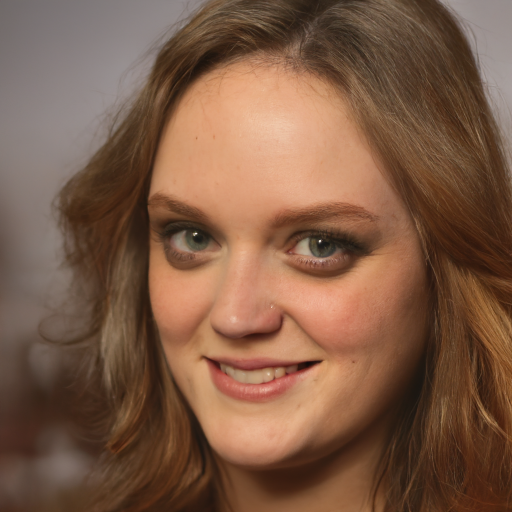}~

    &\includegraphics[width=0.13\textwidth]{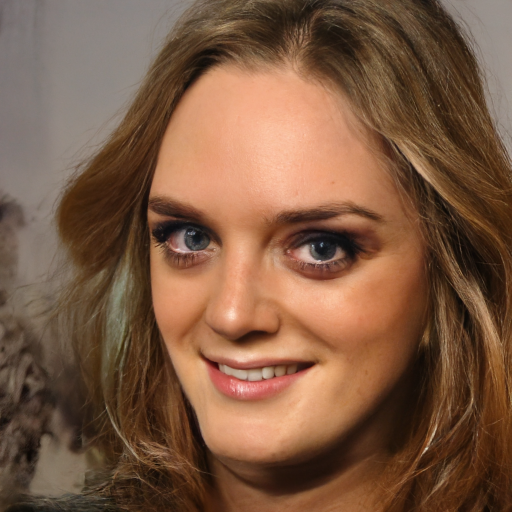}~
        &\includegraphics[width=0.13\textwidth]{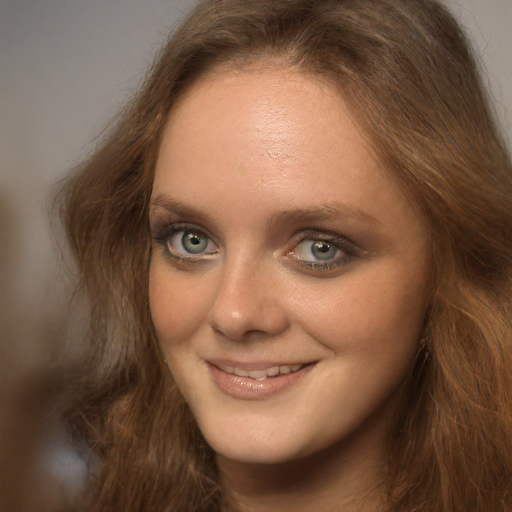}~&\includegraphics[width=0.13\textwidth]{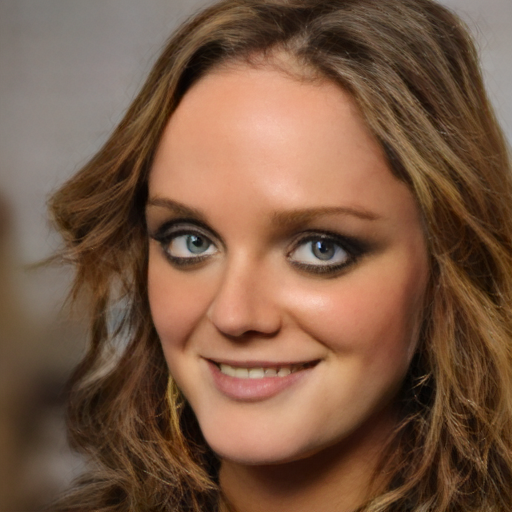}~
		&\includegraphics[width=0.13\textwidth]{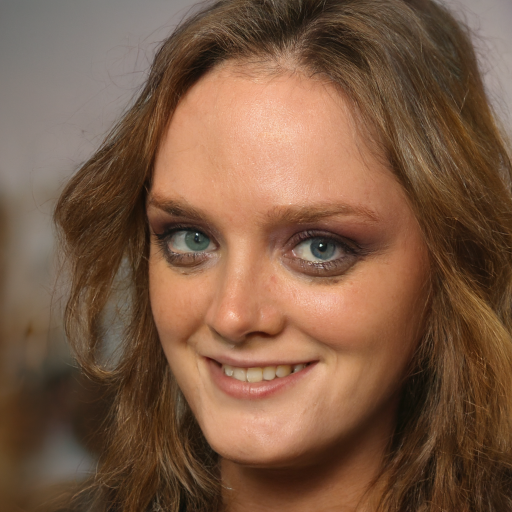}~
		&\includegraphics[width=0.13\textwidth]{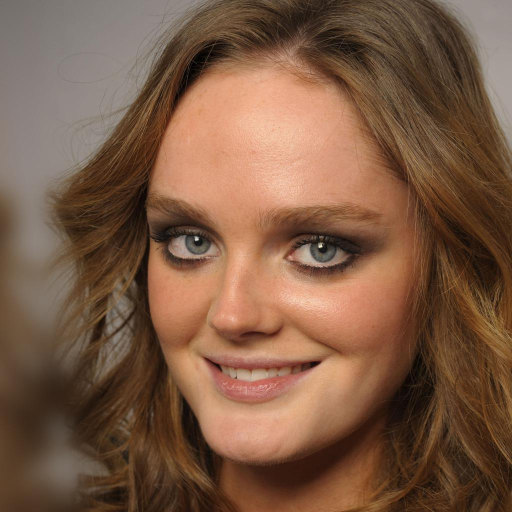}\\
        &\includegraphics[width=0.13\textwidth]{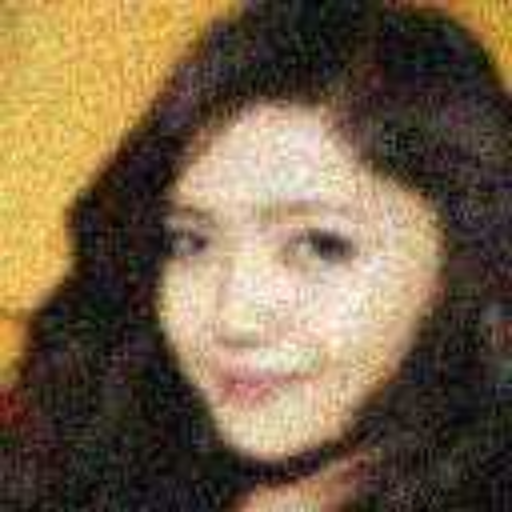}~
		&\includegraphics[width=0.13\textwidth]{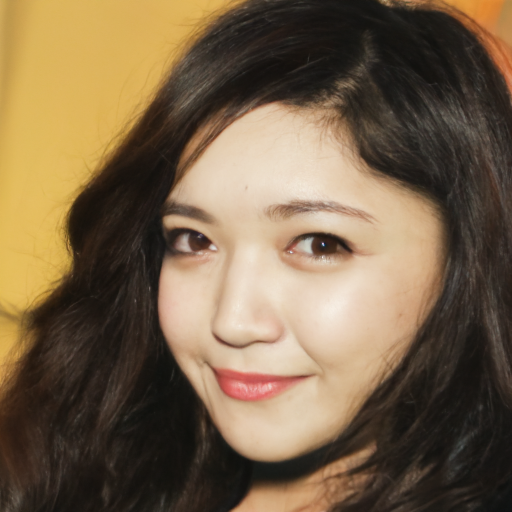}~

    &\includegraphics[width=0.13\textwidth]{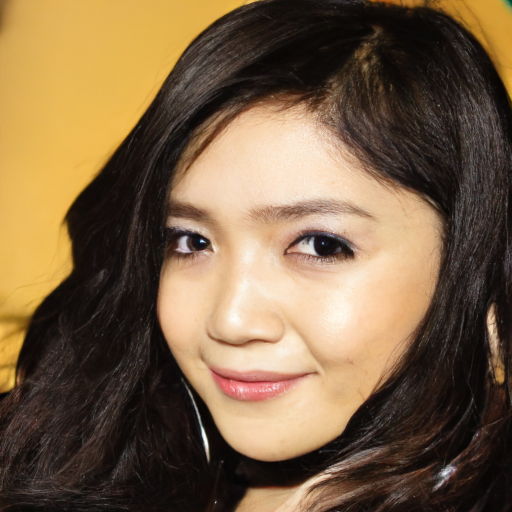}~
        &\includegraphics[width=0.13\textwidth]{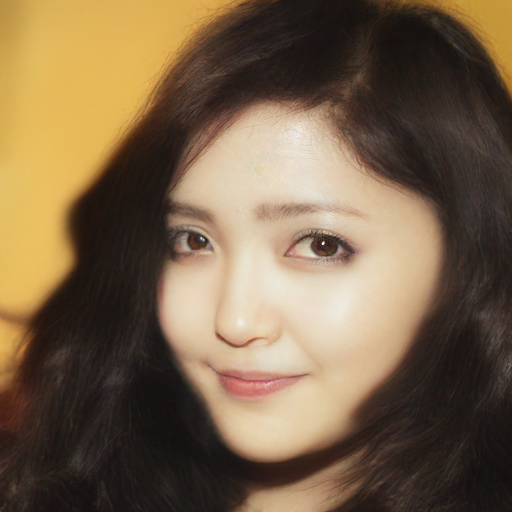}~
        &\includegraphics[width=0.13\textwidth]{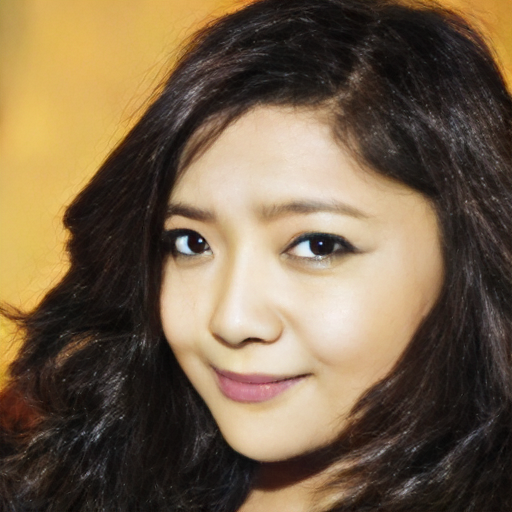
        }~
		&\includegraphics[width=0.13\textwidth]{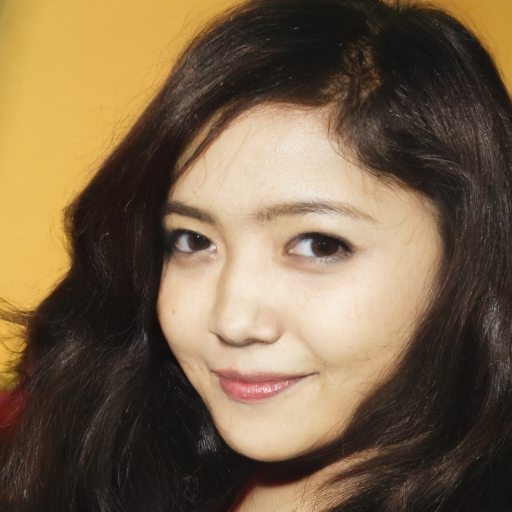}~
		&\includegraphics[width=0.13\textwidth]{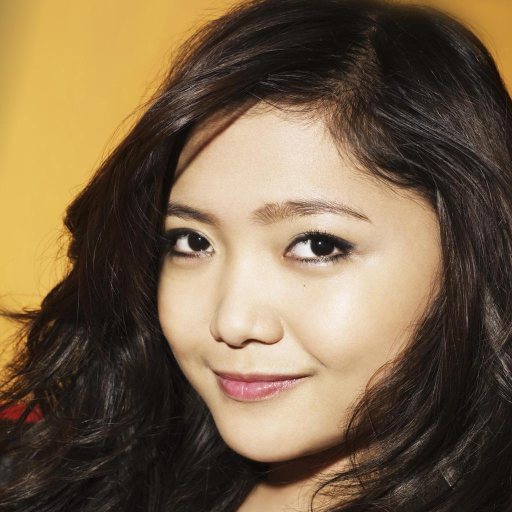}\\
& (a) Input  & (b) DifFace  & (c) PGDiff   & (d) DR2  & (e) {StableSR}& \textbf{(f) Ours} & \textbf{(g) GT}\\
    
	\end{tabular}
		\caption{{Comparisons on 4x \textbf{blind} SR with medium degradation on CelebA-HQ.}} 
	\label{fig:medium}
\end{figure*}

\begin{figure*}[!tp]\footnotesize 
	\centering
\hspace{-0.2cm}
\begin{tabular}{c@{\extracolsep{0.2em}}c@{\extracolsep{0.2em}}c@{\extracolsep{0.2em}}c@{\extracolsep{0.2em}}c@{\extracolsep{0.2em}}c@{\extracolsep{0.2em}}c@{\extracolsep{0.2em}}c@{\extracolsep{0.2em}}c@{\extracolsep{0.2em}}c}
  &\includegraphics[width=0.13\textwidth]{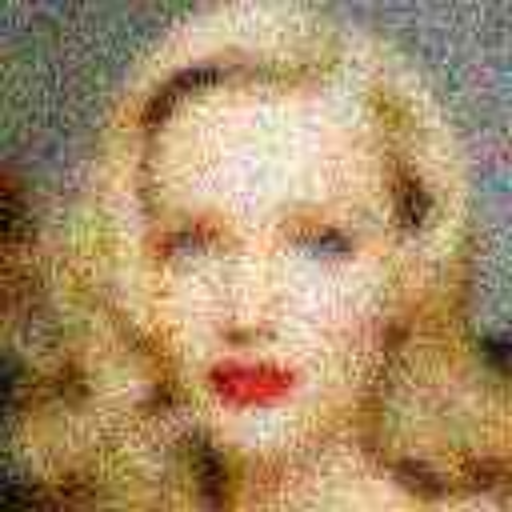}~
		&\includegraphics[width=0.13\textwidth]{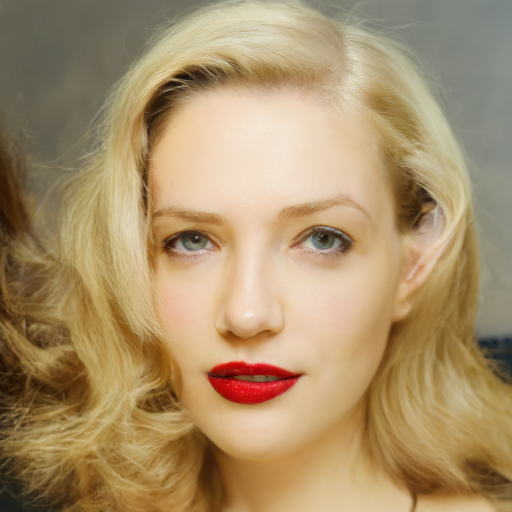}~

    &\includegraphics[width=0.13\textwidth]{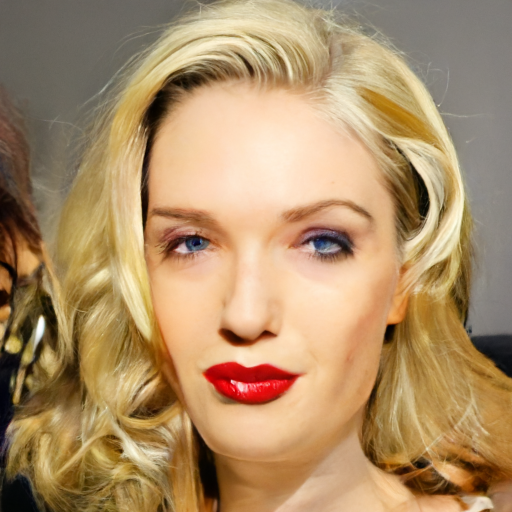}~
        &\includegraphics[width=0.13\textwidth]{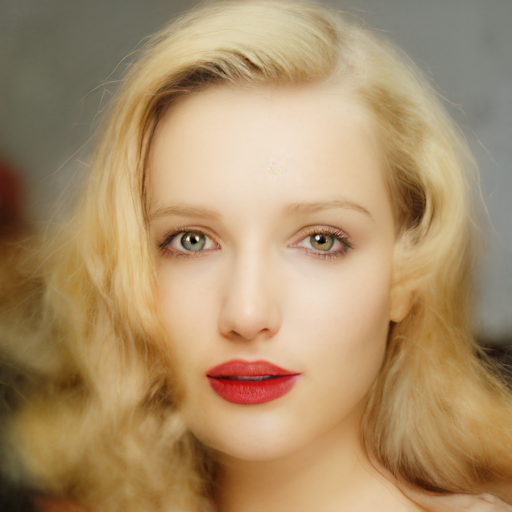}~
                &\includegraphics[width=0.13\textwidth]{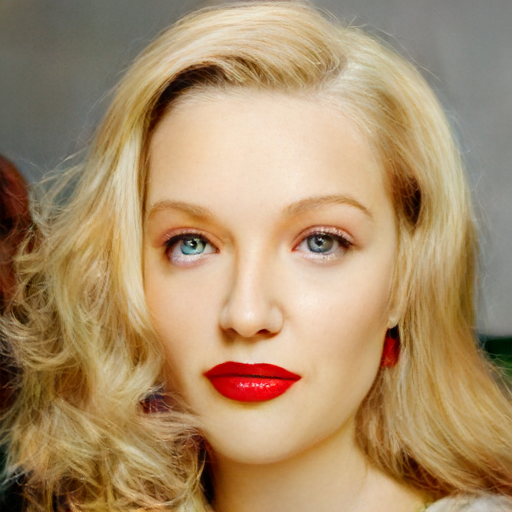}~
		&\includegraphics[width=0.13\textwidth]{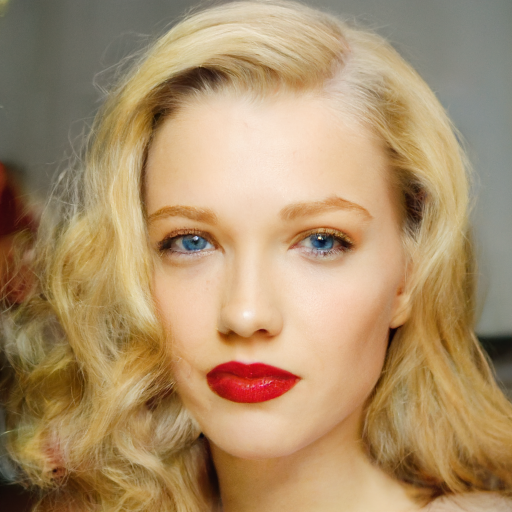}~
		&\includegraphics[width=0.13\textwidth]{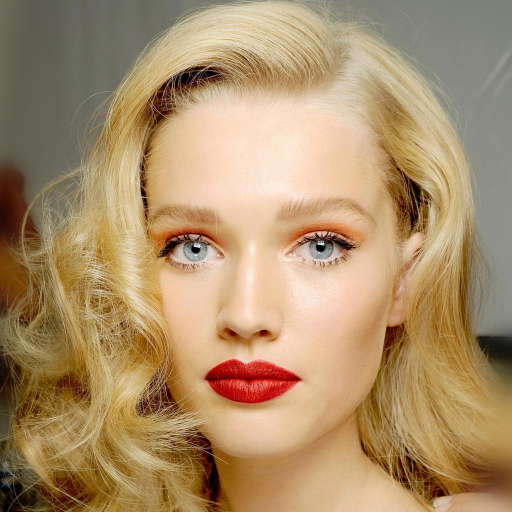}\\
            &\includegraphics[width=0.13\textwidth]{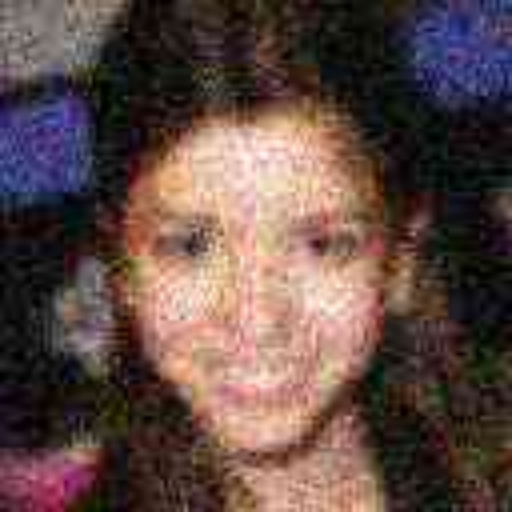}~
		&\includegraphics[width=0.13\textwidth]{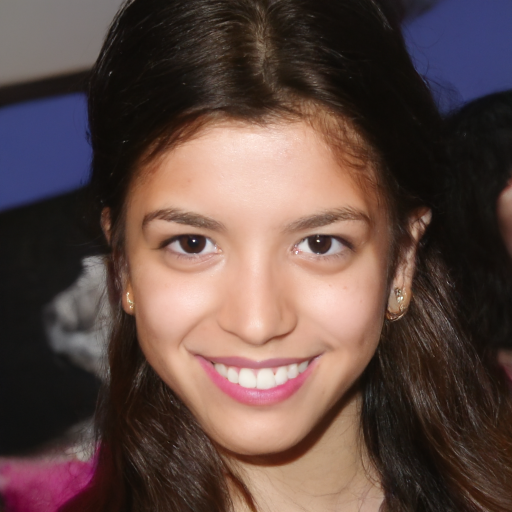}~

    &\includegraphics[width=0.13\textwidth]{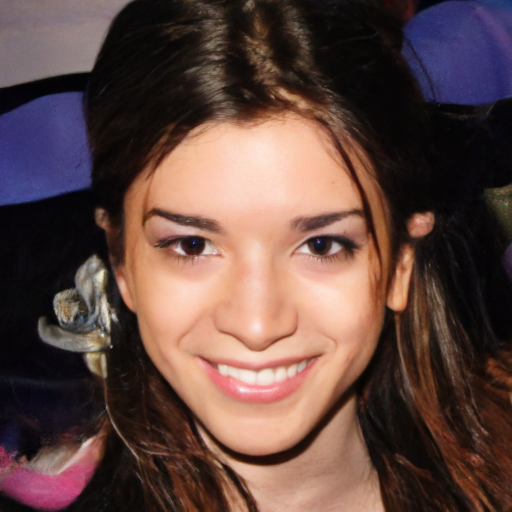}~
        &\includegraphics[width=0.13\textwidth]{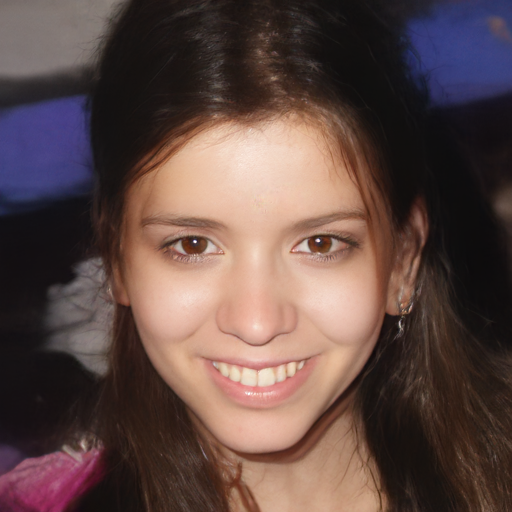}~
        &\includegraphics[width=0.13\textwidth]{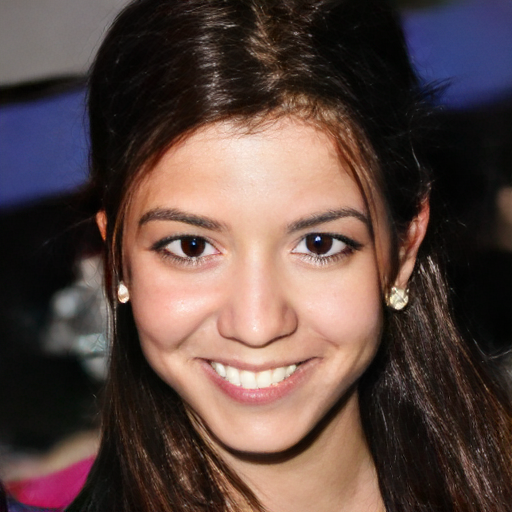}~
		&\includegraphics[width=0.13\textwidth]{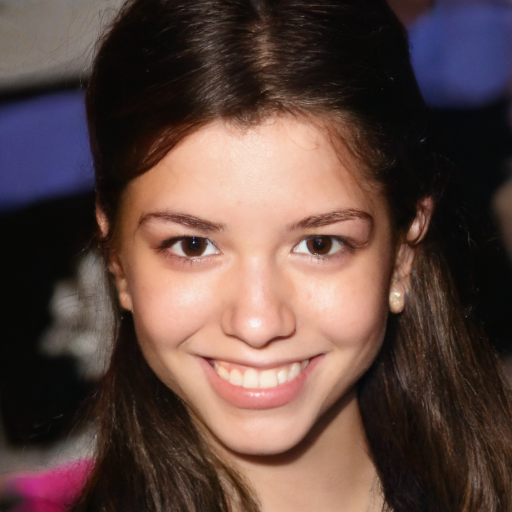}~
		&\includegraphics[width=0.13\textwidth]{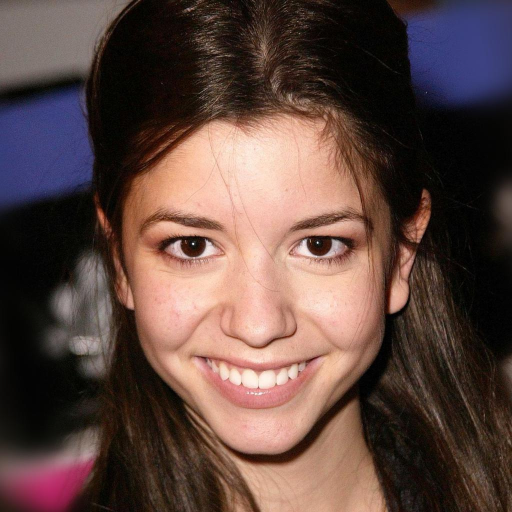}\\
& (a) Input  & (b) DifFace  & (c) PGDiff   & (d) DR2  & (e) {StableSR}& \textbf{(f) Ours} & \textbf{(g) GT}\\
	\end{tabular}
	\caption{{Comparisons on 4x \textbf{blind} SR with severe degradation on CelebA-HQ.}} 
	\label{fig:severe}
\end{figure*}

\subsubsection{Results with Bicubic Downsampling Degradation Model}
\label{sr}
We compare our method with 3 state-of-the-art methods based on diffusion models: ILVR \cite{ilvr}, DDRM \cite{DDRM}, and DPS \cite{dps}. 
As shown in {\color{black}{Table \ref{tab:nonblind}, we evaluate all methods on the problem of bicubic downsampling (4×) with different levels of Gaussian noise on the FFHQ dataset. {Please note that in the first setting which we named `ours', we apply our consensus strategy (see Appendix \ref{appendix}). To further accelerate the sampling strategy, we use DDIM \cite{ddim} as the sampling strategy with time step 250 and present the results as `ours-DDIM'. Further discussion on sampling with DDIM can be seen in Section \ref{ddim_discuss}.}
One can observe that our method outperforms all baseline methods in all metrics. 
Moreover, as it can be seen in Fig. \ref{fig:com}, our algorithm produces high-quality reconstruction results while preserving realistic details. }}

\subsubsection{Results with Non-linear Degradation Model}
\label{complex degradation model}

As described in the introduction, our flexible INN can handle a variety of degradation processes.
In this setting, a bicubic downsampling operator is first performed on a high-quality image and then a JPEG compression degradation is applied to save the low-resolution image into JPEG format, where the jpeg factor and downsampling scale are 20 and 4, respectively.
As shown in Fig. \ref{fig:jpeg10}, our algorithm can still produce high-quality images from heavily degraded input measurements. It can be seen that our results produce realistic details while ensuring data consistency. Note that the methods we compared against in Sec. \ref{sr} are not able to support this type of degradation, so for this case, we only show our results.

\subsubsection{Results with Real Degradation Model}
In Fig. \ref{fig:real}, we show the results of our algorithm on reconstructing images from real-world degradation processes using DRealSR \cite{drealsr} with scale factor 4. 
As for Sec. \ref{complex degradation model}, since we simulate the degradation with INN, our solution is no longer limited by the requirement of knowing the exact expression of the degradation process. 
We can observe that our algorithm can produce, also in this case, high-quality results for real-world degradation.

\subsection{Results on Blind Image Restoration}
\subsubsection{Implementation Details}
Empirically, we set $T$=1000, $N$=400, $\zeta$=1.5, and $l$=1e-5 in Algorithm~\ref{algo:blind}.
Based on the framework for the non-blind setting, we adopt CResblocks \cite{cresmd} for our PNet/UNet, where the number of feature channels is set to 32 and the dimension of $\bm{\gamma}_{deg}$ is 128.   The total number of learnable parameters of our INN is 0.91M. 
We directly utilize the pre-trained implicit degradation estimator in\cite{kdsr} as DEM and we follow DifFace\cite{yue2022difface} to implement our IPN by using a pre-trained SwinIR \cite{liang2021swinir} Network.

We test our method on blind face restoration and blind natural image restoration. For the face photo reconstruction, we utilize a pre-trained unconditional diffusion model trained on the FFHQ training dataset by \cite{yue2022difface} and select 10k images from the FFHQ training dataset to train our INN.
For the natural image reconstruction on ImageNet validation dataset \cite{Imagenet}, we utilize the pre-trained unconditional diffusion model trained on ImageNet \cite{Imagenet} by \cite{guided_diffusion} and use DIV2K \cite{div2k} training dataset to train our INN. 
Following previous works \cite{SPAR,GPEN,GFPGAN,CF,VQFR,wang2023dr2}, we adopt a commonly used degradation model as follows to synthesize training data:
\begin{equation}\label{equ:degradation}
	\bm{y} = \left[(\bm{x}
 \circledast 
 \bm{k}_{\sigma})_{\downarrow_{r}} + \bm{n}_{\delta}\right]_{\mathtt{JPEG}_{q}},
\end{equation}
where the high quality image $x$ is first convolved with a Gaussian blurring kernel $k_{\sigma}$ followed by a downsampling operation with a scale factor $r$. After that, additive white Gaussian noise $n_{\delta}$ is added to the image, and finally the noisy image is compressed by JPEG with quality factor $q$. To train our INN, we set $r$ = 4, and randomly sample values of $\sigma$, $\delta$ and $q$ from the intervals [3,9], [5,50], and [30,80] respectively.
The reconstruction results are evaluated with PSNR, LPIPS \cite{lpips}, BRISQUE \cite{brisque},  Identity Similarity (IDS) using the ArcFace similarity~\cite{arcface}.

\begin{table}[]
    \caption{Complexity comparison with state-of-the-art blind image restoration approaches (Runtime/Params/Iterations).}
    \label{tab:runtime}
    \centering
    \begin{tabular}{|l||c|c|c|c|c|}
\hline 
& DR2&DifFace  & PGDiff & StableSR  & Ours\\ \hline \hline 

Runtime                     &  0.96s& 3.36s
   &59.37s &126.07s&43.85s \\  
   Params                     &86M  & 16M
   & 17M & 94M+105M & 16M+1M \\  
   Iterations                     &100  & 100
   & 1000 & 1000& 400\\

\hline  
  
    \end{tabular}
\end{table}
\begin{table}[]
    \caption{Quantitative (BRISQUE/Identity Similarity (IDS)) comparison on real-world face image restoration. \textbf{Bold} and \underline{underlined} texts represent the best and second best performance.}
    \label{tab:realface}
    \centering
    \begin{tabular}{|l||C{1.8cm} |C{1.8cm}|}
\hline 
             Methods                            & Brisque $ \downarrow$ &IDS $ \uparrow$ \\ \hline \hline 

PGDiff~\cite{yang2023pgdiff}                      & \underline{21.82} & 0.8013            \\  
Difface~\cite{yue2022difface}                & 25.96& 0.7697          \\  
DR2~\cite{wang2023dr2}                   &25.01& \textbf{0.9503}         \\
StableSR~\cite{wang2023dr2}                   &25.30& 0.9302        \\

Ours                   &\textbf{20.35}& \underline{0.9315}       \\
\hline  

    \end{tabular}
\end{table}

\subsubsection{Results on Synthetic Degradation}
\begin{figure*}[!tp]\footnotesize 
	\centering
\hspace{-0.2cm}
\begin{tabular}{c@{\extracolsep{0.2em}}c@{\extracolsep{0.2em}}c@{\extracolsep{0.2em}}c@{\extracolsep{0.2em}}c@{\extracolsep{0.2em}}c@{\extracolsep{0.2em}}c@{\extracolsep{0.2em}}c@{\extracolsep{0.2em}}c}
  &\includegraphics[width=0.13\textwidth]{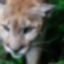}~
        &\includegraphics[width=0.13\textwidth]{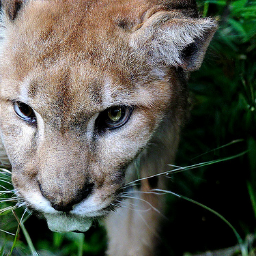}~
               &\includegraphics[width=0.13\textwidth]{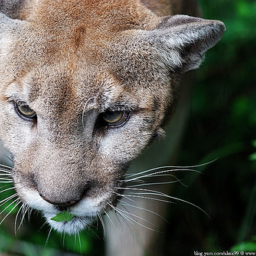}~
         &\includegraphics[width=0.13\textwidth]{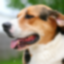}~
        &\includegraphics[width=0.13\textwidth]{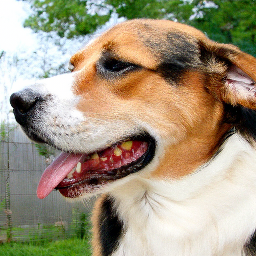}~
               &\includegraphics[width=0.13\textwidth]{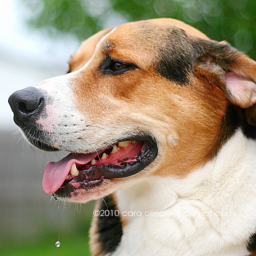}~\\
                                                        &\includegraphics[width=0.13\textwidth]
                              {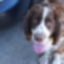}~
        &\includegraphics[width=0.13\textwidth]{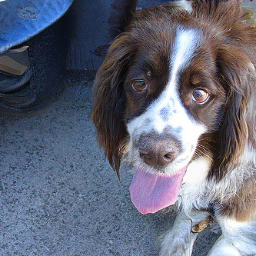}~
               &\includegraphics[width=0.13\textwidth]{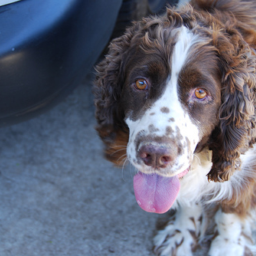}~
       &\includegraphics[width=0.13\textwidth]{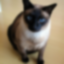}~
        &\includegraphics[width=0.13\textwidth]{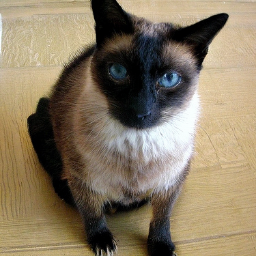}~
               &\includegraphics[width=0.13\textwidth]{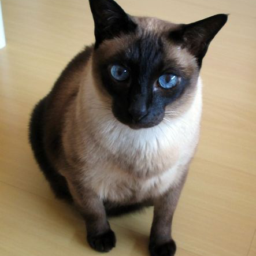}~\\
    & Input  & \textbf{Ours}   & Ground-truth   &Input  &{\textbf{Ours}} &Ground-truth\\
	\end{tabular}
	\caption{{Result of our BlindINDIGO on 4x super-resolution on the ImageNet dataset.}} 
	\label{fig:srimagenet}
\end{figure*}

We test our method on CelebA HQ 512×512 1k validation dataset \cite{celeba} and ImageNet validation dataset \cite{Imagenet} on synthetic degradation.
We compare our method with 4 state-of-the-art methods based on diffusion models: PGDiff \cite{yang2023pgdiff}, DifFace \cite{yue2022difface}, DR2 \cite{wang2023dr2}, and StableSR \cite{stablesr}. To evaluate these methods on different levels of degradation, we test them on mild ($\sigma$=4, $\delta$=15, $q$=70), medium ($\sigma$=6, $\delta$=25, $q$=50)  and severe ($\sigma$=8, $\delta$=35, $q$=30) degradations respectively. We provide the quantitative comparison on different levels of degradations in Table \ref{tab:blind} and we see that our approach achieves significant gains over existing works, in particular, for medium and severe degradations. {To further accelerate the sampling strategy, we use DDIM \cite{ddim} as the sampling strategy with T=250, N=100 and present the results as `ours-DDIM'. One can observe that DDIM can speed up the reconstruction process with a slight reduction in perceptual quality. Further discussion on sampling with DDIM can be seen in Section \ref{ddim_discuss}.} Qualitative comparisons in Fig. \ref{fig:mild}, Fig. \ref{fig:medium}, and Fig. \ref{fig:severe} demonstrate the superiority of our BlindINDIGO in comparison to existing methods on different levels of degradations. {\color{black}{Fig. \ref{fig:srimagenet} demonstrates the robustness of our approach, enabling its application to a variety of categories, including cats, dogs, and lions.}} Here, we measure the runtime of all the approaches on an Nvidia RTX 3080 GPU and show complexity comparison in Table \ref{tab:runtime}. The parameter numbers here do not include the parameters of the pre-trained diffusion model.

\subsubsection{Results with Real Degradation Model} 
\begin{figure*}[!tp]\footnotesize 
	\centering
\hspace{-0.2cm}
\begin{tabular}{c@{\extracolsep{0.2em}}c@{\extracolsep{0.2em}}c@{\extracolsep{0.2em}}c@{\extracolsep{0.2em}}c@{\extracolsep{0.2em}}c@{\extracolsep{0.2em}}c@{\extracolsep{0.2em}}c@{\extracolsep{0.2em}}c}
      &\includegraphics[width=0.15\textwidth]{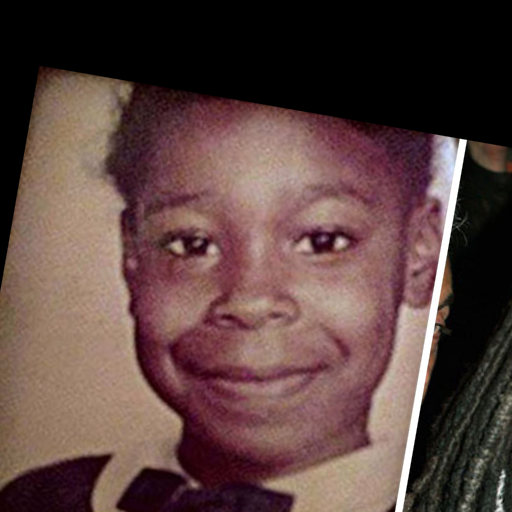}~
		&\includegraphics[width=0.15\textwidth]{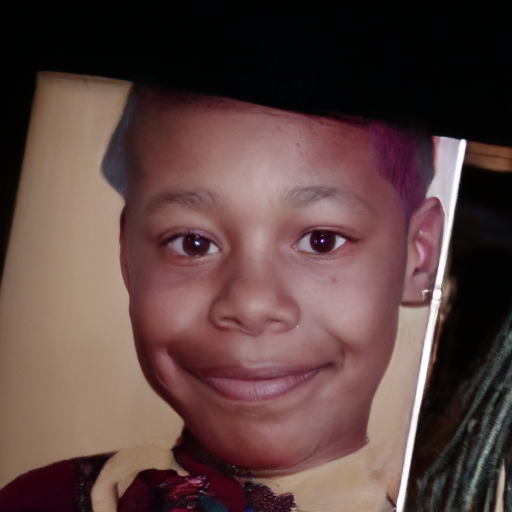}~
    &\includegraphics[width=0.15\textwidth]{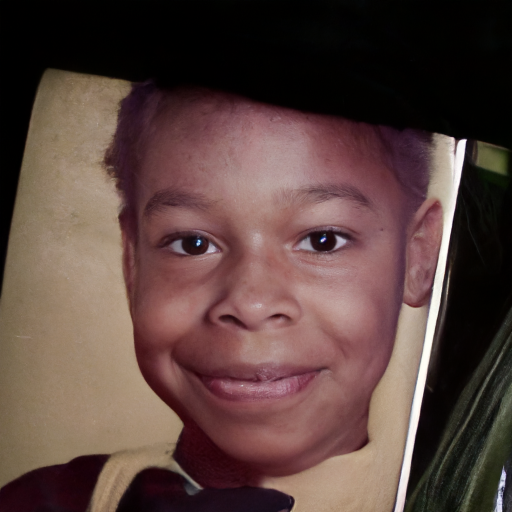}~
        &\includegraphics[width=0.15\textwidth]{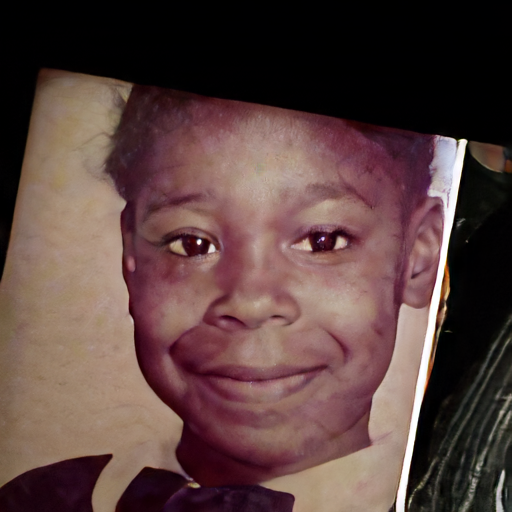}~
		&\includegraphics[width=0.15\textwidth]{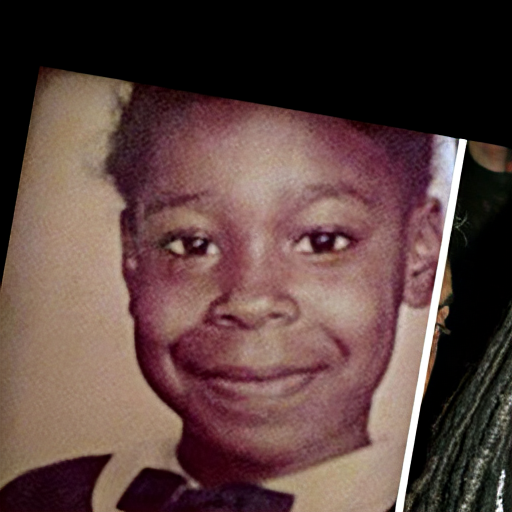}~
		&\includegraphics[width=0.15\textwidth]{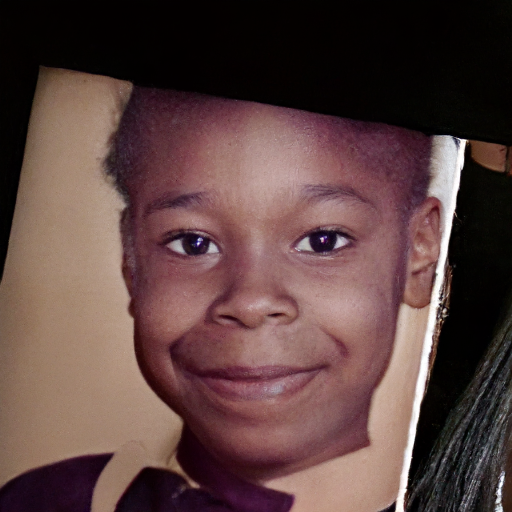}\\
        &\includegraphics[width=0.15\textwidth]{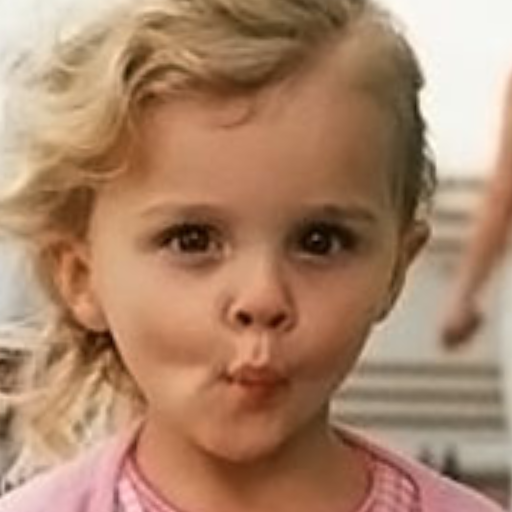}~
		&\includegraphics[width=0.15\textwidth]{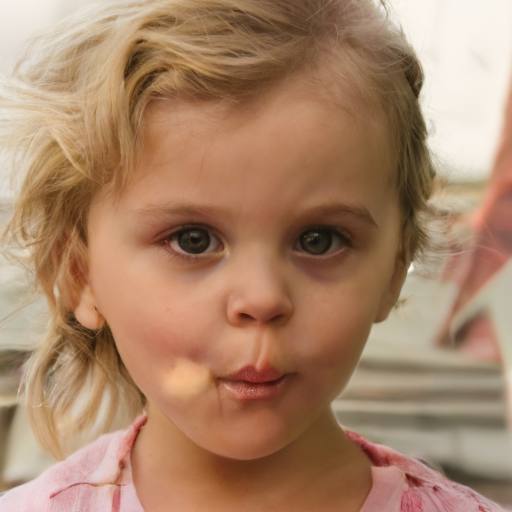}~
    &\includegraphics[width=0.15\textwidth]{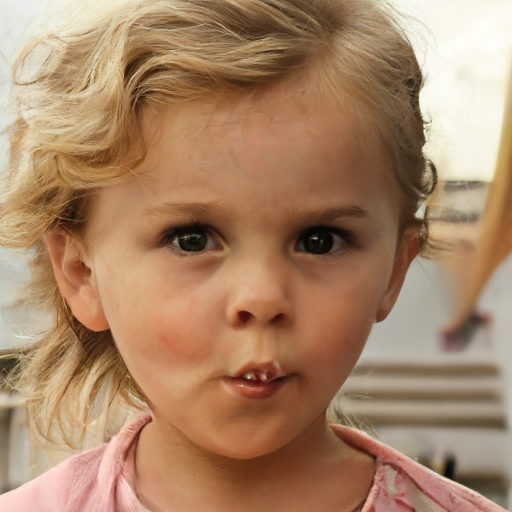}~
        &\includegraphics[width=0.15\textwidth]{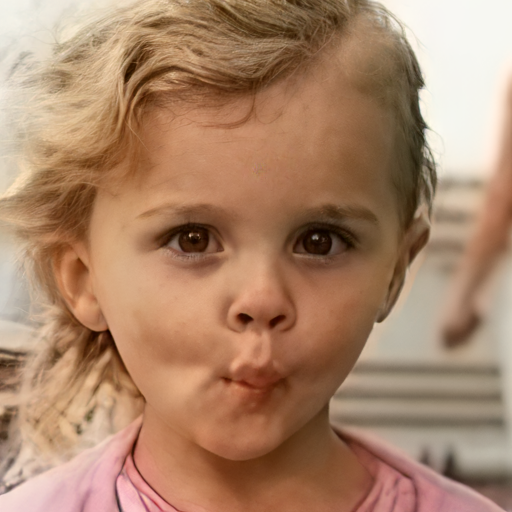}~
		&\includegraphics[width=0.15\textwidth]{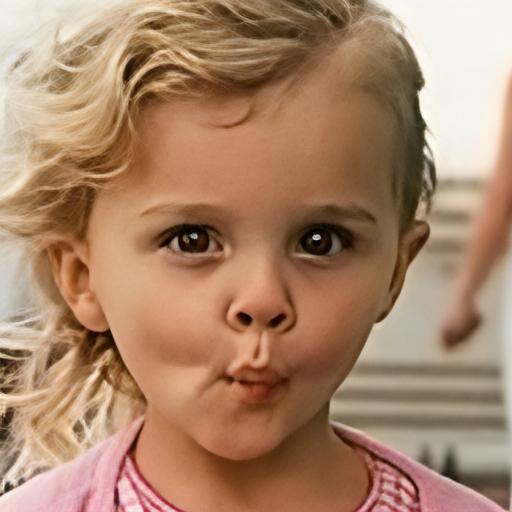}~
		&\includegraphics[width=0.15\textwidth]{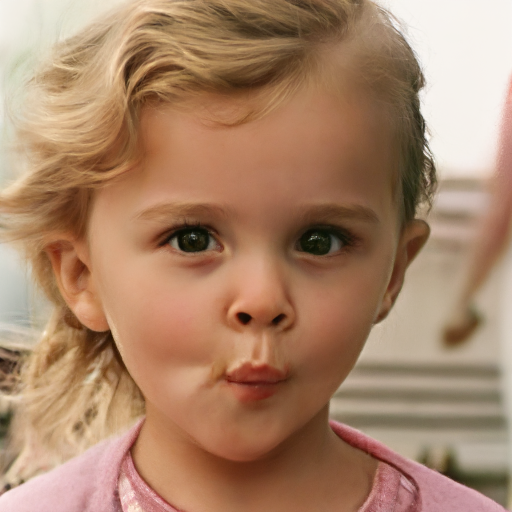}\\
      
    & Input  & DifFace  & PGDiff   &DR2   &StableSR  &\textbf{Ours} \\
	\end{tabular}
    \vspace{-0.1cm}
	\caption{Comparisons with blind state-of-the-art image restoration approaches  \cite{wang2023dr2,yue2022difface,yang2023pgdiff,stablesr} on real-world dataset Celaba-Child.} 
	\label{fig:wider}
\end{figure*}

We also apply our BlindINDIGO on a real-world dataset
CelebChild \cite{GFPGAN} for evaluating the generalization of the proposed method. 
CelebChild-Test contains 180 child faces of celebrities collected from {the Internet}. They are low-quality and many of them are black-and-white old photos. Since no ground-truth images are available for this setting, we compare the image quality using BRISQUE\cite{brisque} and compare the image fidelity with Identity Similarity (IDS) using the ArcFace similarity~\cite{arcface} 
in Table \ref{tab:realface}. We observe that our approach achieves best or second best scores, demonstrating its superiority in the generation of high-quality images and effectiveness in preserving identity. The qualitative comparison on 
CelebChild\cite{GFPGAN} is shown in Fig. \ref{fig:wider}. 
\begin{figure}[!tp]\footnotesize 
	\centering 
\begin{subfigure}[b]{0.48\textwidth}
	\centering 
\begin{tabular}{c@{\extracolsep{0.em}}c@{\extracolsep{0.em}}c@{\extracolsep{0.em}}c@{\extracolsep{0.em}}c@{\extracolsep{0.em}}c@{\extracolsep{0.em}}c@{\extracolsep{0.em}}c}
          &\includegraphics[width=0.18\textwidth]{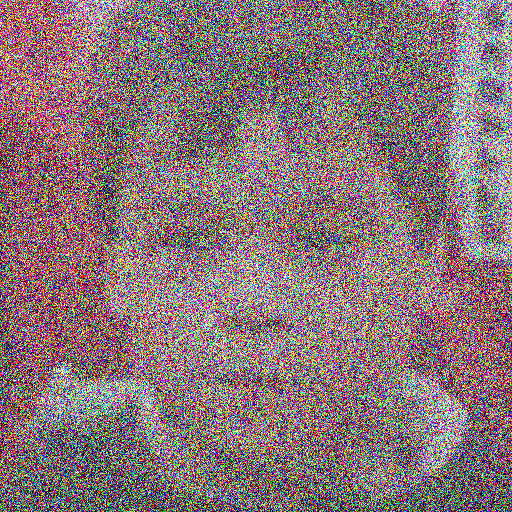}~ &\includegraphics[width=0.18\textwidth]{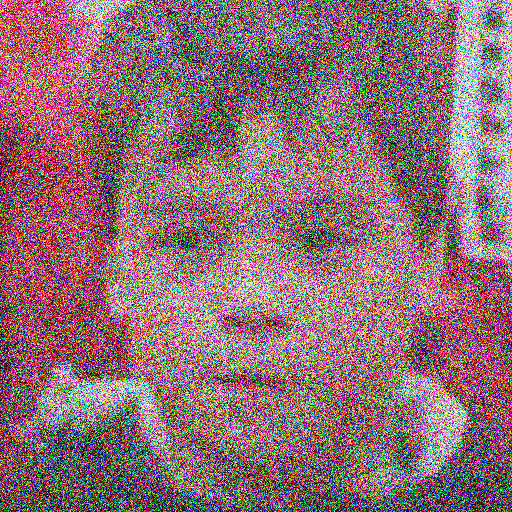}~ &\includegraphics[width=0.18\textwidth]{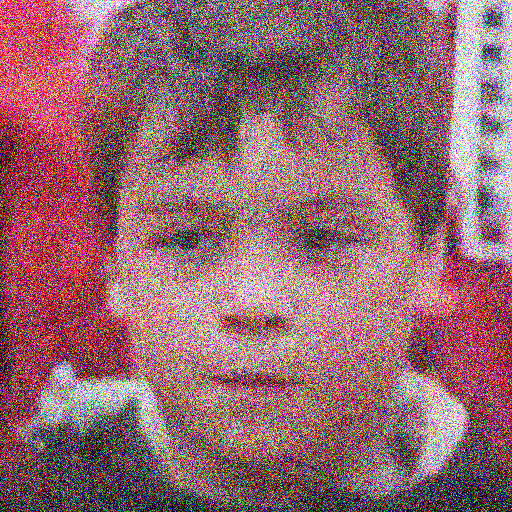}~                &\includegraphics[width=0.18\textwidth]{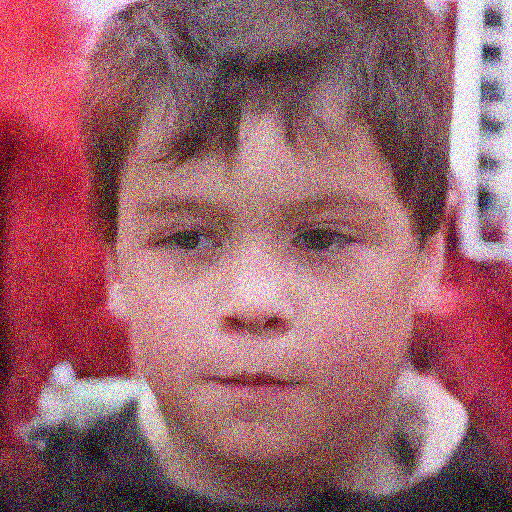}~ &\includegraphics[width=0.18\textwidth]{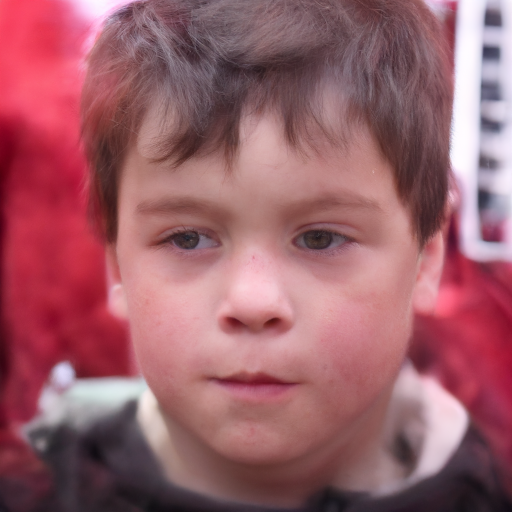}~\\
             & $\bm{x}_{400}$  &  $\bm{x}_{300}$  &   $\bm{x}_{200}$&  $\bm{x}_{100}$ &  $\bm{x}_{0}$  \\
	\end{tabular}
	\caption{Noisy intermediate results.} 
	\label{fig:visual_result}
\end{subfigure}
 \vspace{0.8cm}
 \hfill
\begin{subfigure}[b]{0.48\textwidth}
	\centering
\vspace{0.3cm}
\begin{tabular}{c@{\extracolsep{0.1em}}c@{\extracolsep{0.1em}}c@{\extracolsep{0.1em}}c}
		&\includegraphics[width=0.3\textwidth]{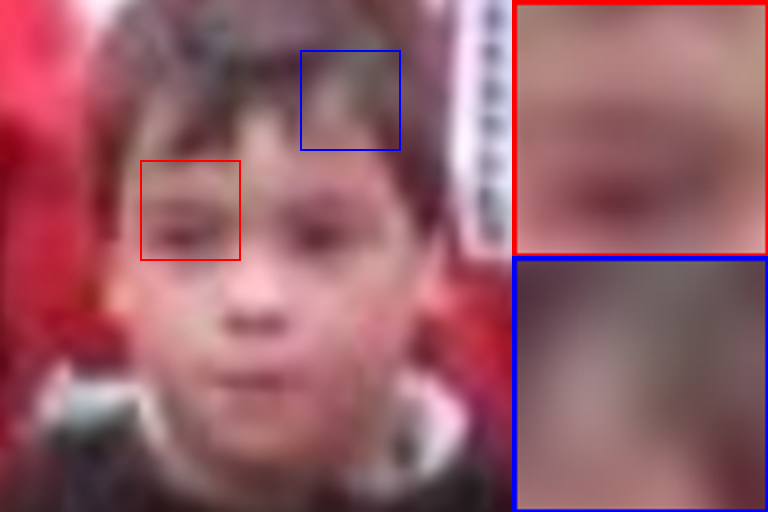}~
&\includegraphics[width=0.3\textwidth]{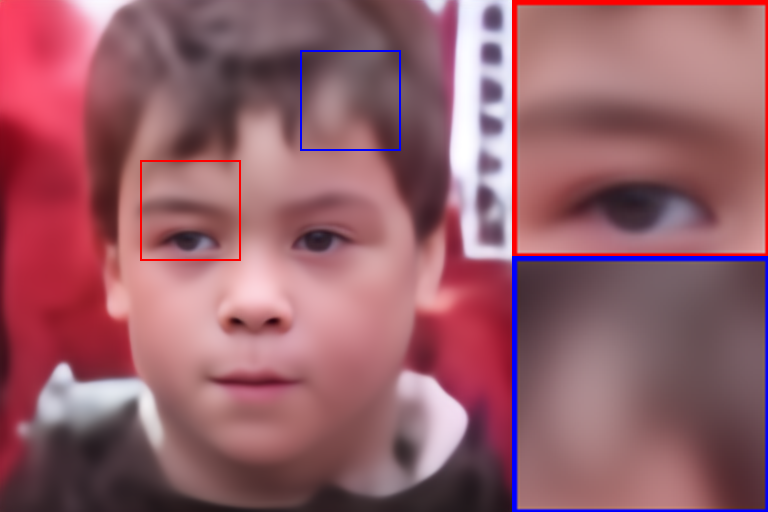}~
		&\includegraphics[width=0.3\textwidth]{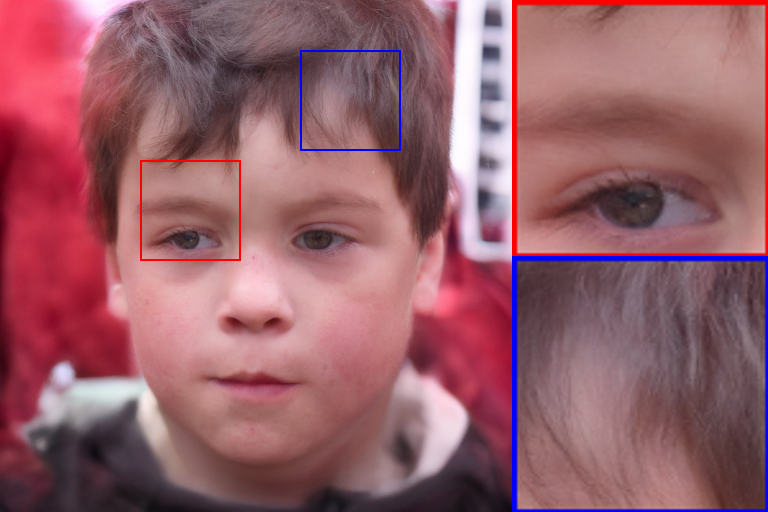}~\\
& Low-Resolution  & $\bm{y}_0$  & $\bm{x}_0$   \\
     
		&\includegraphics[width=0.3\textwidth]{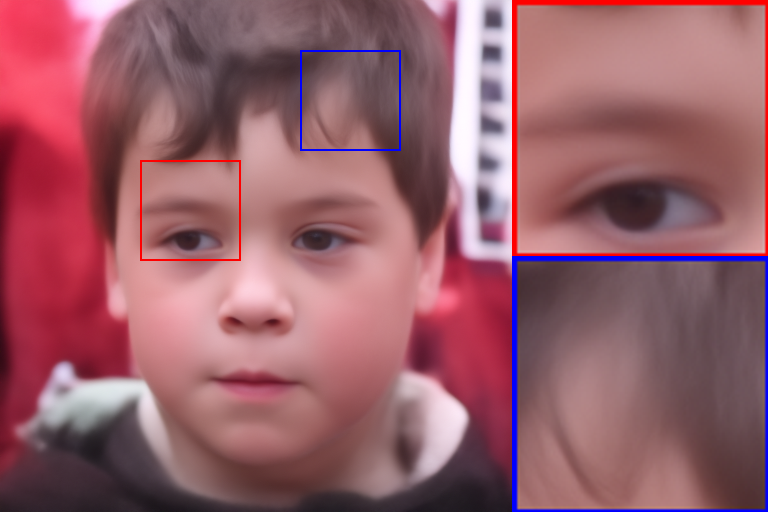}~

    &\includegraphics[width=0.3\textwidth]{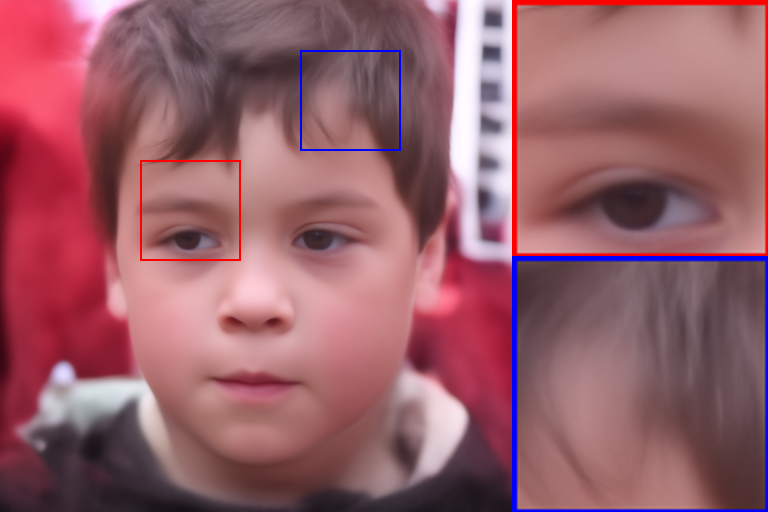}~
        &\includegraphics[width=0.3\textwidth]{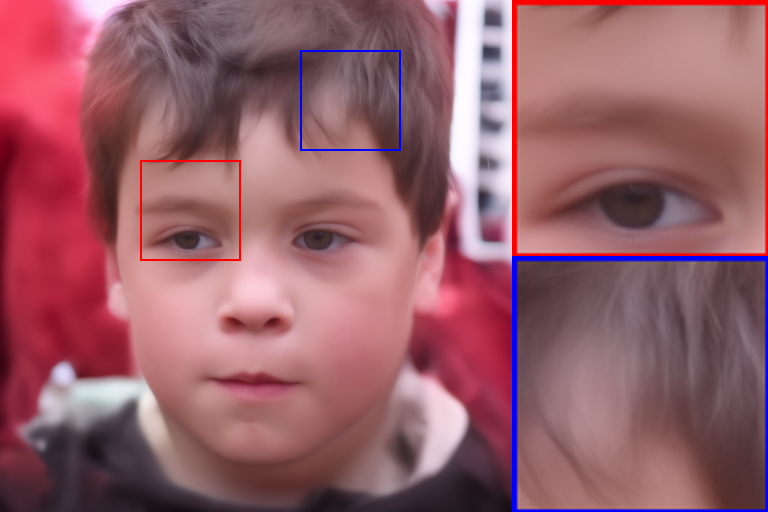}~\\
       & $\bm{x}_{0,300}$  & $\bm{x}_{0,200}$  & $\bm{x}_{0,100}$   \\

	\end{tabular}
    \vspace{-0.1cm}
	\caption{{Clean intermediate results.}} 
 \vspace{-1cm}
	\label{fig:intermediate}
\end{subfigure}
\caption{{Sampling process of our approach.}} 
\label{intermediate}
\end{figure}

\subsection{Analysis and Discussion}
{\subsubsection{Comparison to conditional diffusion model and standard supervised learning approach}
Different from training a conditional diffusion model for a specific inverse problem from scratch, we keep the 
same pre-trained
diffusion model and only modify the inference procedure with the guidance of INN to enable sampling from a conditional distribution.
This strategy can efficiently leverage the trained diffusion model trained on huge amount of data to make it serve as a strong generative prior in different inverse problems. In addition, it saves the cost of training, since we only need to train the INN (0.71 M). 
Furthermore, our blind INDIGO does not necessarily need to have access to labeled datasets, while conditional diffusion models need them. In this section, we compare our approach with two representative methods from conditional diffusion models: I2SB \cite{liu20232} and standard supervised learning approaches SwinIR \cite{liang2021swinir} on 4x SR with different levels of Gaussian noise on Imagenet validation dataset.  Table~\ref{tab:i2sb} shows that our approach achieves competitive reconstruction results in the noiseless case and outperforms significantly the other methods in noisy settings.}

\begin{table*}[]
    \caption{{Quantitative (PSNR$\uparrow$/FID$\downarrow$) comparison on 4${\times}$ SR with different levels of Gaussian noise. \textbf{Bold} and \underline{underlined} texts represent the best and second best performance.} }
    \label{tab:i2sb}
    \centering
        \begin{tabular}{|l| |c|c||c|c||c|c||c|}
\hline 
            \multirow{2}{*}{Methods }                              & \multicolumn{2}{c||}{$\sigma$=0}& \multicolumn{2}{c||}{$\sigma$=0.05} &  \multicolumn{2}{c||}{$\sigma$=0.10} &\multirow{2}{*}{Time}  \\ \cline{2-7}                             & PSNR$\uparrow$    & FID$\downarrow$ &  PSNR$\uparrow$   & FID$\downarrow$    & PSNR $\uparrow$   & FID$\downarrow$&  \\ \hline \hline 
{{SwinIR} }      &{24.60}&118.55&   \underline{{{24.29}}} & \underline{{125.41}}   & \underline{23.51} &\underline{{152.07 }}   & 0.11s      \\  
I2SB           &\underline{25.46} &\textbf{55.27}&   21.90&138.53   & 17.62 &248.15  & 36.20s         \\  
 \textbf{Ours}                 & \textbf{{26.28}}&{\underline{79.98}}       & \textbf{25.70}&{\textbf{81.85}} 
& {\textbf{24.58}}&{\textbf{99.08}} & 54.27s \\  
\hline

    \end{tabular}
\end{table*}
\subsubsection{Analysis on the sampling process of our approach} 
We show the visual results of the sampling process in Fig. \ref{intermediate}. To clearly illustrate the comparison, we present the noisy results during the sampling process in Fig. \ref{intermediate}(a), and the clean results in Fig. \ref{intermediate}(b). 

Firstly, as shown in Algorithm~\ref{algo:blind}, given a degraded image $\bm{y}$ during inference, our approach first predicts a clean version $\bm{y}_{0}$ with the Initialization Prediction Network (IPN).
As shown in the first row of Fig. \ref{intermediate}(b), IPN effectively predict a smooth result $\bm{y}_0$ which still lacks details compared to our final result $\bm{x}_{0}$.

Secondly, since we set $T$=1000 and $N$=400 in Algorithm~\ref{algo:blind}, we start from $\bm{x}_{400}$ instead of $\bm{x}_{1000}$. As shown in Fig. \ref{intermediate} (a), the noise gradually decreases with iterations from $t$=400 to 0. And the second row of Fig. \ref{intermediate} (b) shows the effectiveness of our data-consistency step guided by the INN.

\subsubsection{Effect of our INN} 
\begin{figure}[!tp]\footnotesize 
	\centering 
 \hfill
\begin{subfigure}[b]{0.48\textwidth}
	\centering
\begin{tabular}{c@{\extracolsep{0.1em}}c@{\extracolsep{0.1em}}c@{\extracolsep{0.1em}}c@{\extracolsep{0.1em}}c}
		&\includegraphics[width=0.22\textwidth]{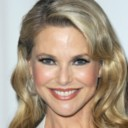}~        &\includegraphics[width=0.22\textwidth]{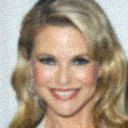}~
&\includegraphics[width=0.22\textwidth]{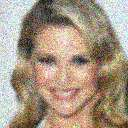}~                &\includegraphics[width=0.22\textwidth]{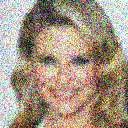}~\\
	\end{tabular}
    \vspace{-0.15cm}
	\caption{Low-resolution inputs with different degradation levels.} 
 \vspace{0.1cm}
	\label{fig:intermediate}
\end{subfigure}
\begin{subfigure}[b]{0.48\textwidth}
	\centering
\begin{tabular}{c@{\extracolsep{0.1em}}c@{\extracolsep{0.1em}}c@{\extracolsep{0.1em}}c@{\extracolsep{0.1em}}c}
 &\includegraphics[width=0.22\textwidth]{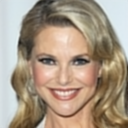}~
        &\includegraphics[width=0.22\textwidth]{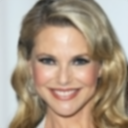}~
		&\includegraphics[width=0.22\textwidth]{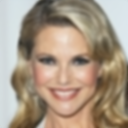}~                &\includegraphics[width=0.22\textwidth]{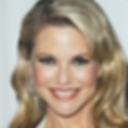}~\\
	\end{tabular}
        \vspace{-0.15cm}
	\caption{Coarse parts generated by forward INN with different conditions.} 
  \vspace{0.1cm}
	\label{fig:intermediate}
\end{subfigure}
\begin{subfigure}[b]{0.48\textwidth}
	\centering
\begin{tabular}{c@{\extracolsep{0.1em}}c@{\extracolsep{0.1em}}c@{\extracolsep{0.1em}}c@{\extracolsep{0.1em}}c}
    &\includegraphics[width=0.22\textwidth]{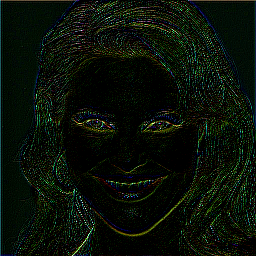}~
        &\includegraphics[width=0.22\textwidth]{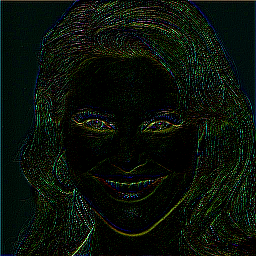}~
		&\includegraphics[width=0.22\textwidth]{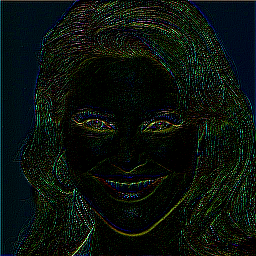}~                &\includegraphics[width=0.22\textwidth]{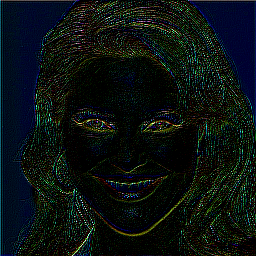}~\\
	\end{tabular}
        \vspace{-0.15cm}
	\caption{Detail parts generated by forward INN with different conditions.} 
	\label{fig:intermediate}
\end{subfigure}
\caption{{Effect of our INN.}} 
\label{inneffect}
\end{figure}

Our INN is designed to simulate the degradation process, as in $[\bm{c},\bm{d}] =f_{\phi}(\bm{x}, \bm{\gamma}_{deg})$. To demonstrate the ability of our INN to simulate different levels of degradation, we present the results of $\bm{c}$ and $\bm{d}$ in the second and third rows of Fig. \ref{inneffect}. As a reference, the first row shows the degraded measurements $\bm{y}$. From left to right, the level of degradation gradually increases, and the coarse part $\bm{c}$ in the second row also follow this trend.

\begin{table}[]
\captionsetup{skip=10pt}
    \caption{{Performance of our approach with different NFE using DDPM and DDIM in terms of PSNR$\uparrow$/ LPIPS$\downarrow$  on 4${\times}$ SR.}}
    \label{tab:nfe}
    \centering
    \begin{tabular}{|c|c||c|c|}
\hline 
             Sampling  & NFE                            & PSNR&LPIPS   \\ \hline \hline 

DDPM&1000 (default)&{28.12} &\textbf{0.0806}    \\  
DDIM & 500 & 28.21 & {{0.0847}}  \\ 
DDIM&250 & \textbf{28.31} & {{0.0914}}  \\ 
DDIM&200 & 28.20 & {{0.0986}}  \\ 
DDIM&100 & 26.01 & {{0.1709}}  \\ 
\hline  

    \end{tabular}
\end{table}
\subsubsection{Effect of step size} 
\begin{figure*}[!tp]\footnotesize 
	\centering
\hspace{-0.26cm}
\begin{tabular}{c@{\extracolsep{0.1em}}c@{\extracolsep{0.1em}}c@{\extracolsep{0.1em}}c@{\extracolsep{0.1em}}c@{\extracolsep{0.1em}}c@{\extracolsep{0.1em}}c@{\extracolsep{0.1em}}c}
		  		&\includegraphics[width=0.13\textwidth]{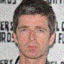}~
      &\includegraphics[width=0.13\textwidth]{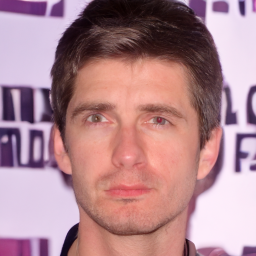}~
		&\includegraphics[width=0.13\textwidth]{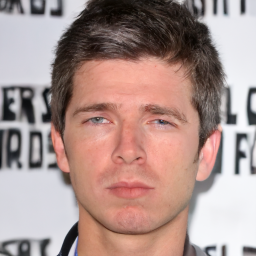}~
        &\includegraphics[width=0.13\textwidth]{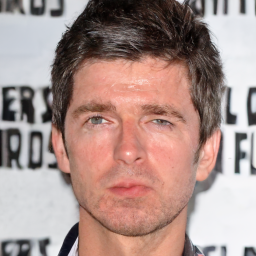}~
		&\includegraphics[width=0.13\textwidth]{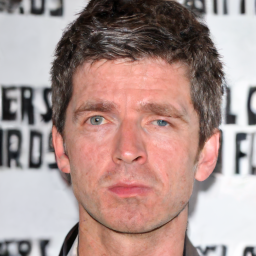}~

      		&\includegraphics[width=0.13\textwidth]{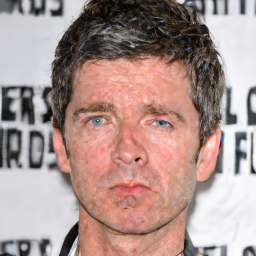}~
		&\includegraphics[width=0.13\textwidth]{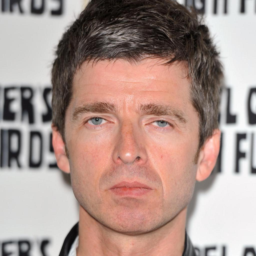}\\
    &\includegraphics[width=0.13\textwidth]{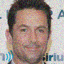}~
            &\includegraphics[width=0.13\textwidth]{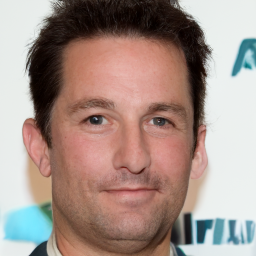}~
          &\includegraphics[width=0.13\textwidth]{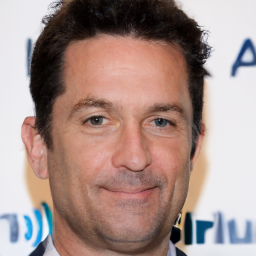}~
		&\includegraphics[width=0.13\textwidth]{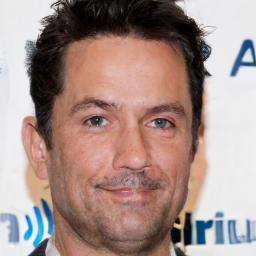}~
  &\includegraphics[width=0.13\textwidth]{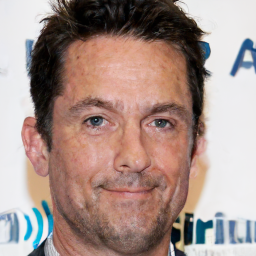}~

  &\includegraphics[width=0.13\textwidth]{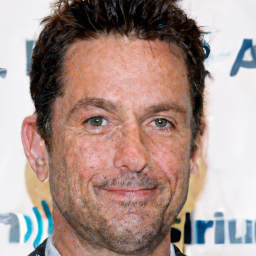}~
		&\includegraphics[width=0.13\textwidth]{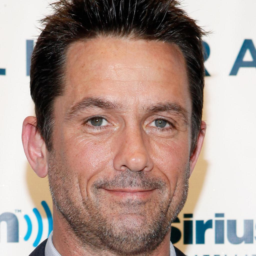}\\
   &Bicubic& $\zeta=0.1$ & $\zeta=0.3$  & $\zeta=0.5$   &$\zeta=0.7$  &$\zeta=1.0$ &Ground Truth\\
	\end{tabular}
	\caption{Ablation study on the choice of step size schedule for our INDIGO.} 
	\label{fig:stepsize}
\end{figure*}

The value of step size $\zeta$ is essentially the weight that is given to the data consistency of the inverse problem. Fig. \ref{fig:stepsize} shows results with different step sizes $\zeta$ in our non-blind INDIGO in the case of noise level $\sigma=0.1$. One can observe that with low values of the step size, the results we obtain have lower consistency with the given measurements. On the other hand, setting the step size value too high leads to artefacts that tend to amplify the noise. Therefore, we set the step size $\zeta=0.5$ and $\zeta=1.5$ by default in our non-blind INDIGO and BlindINDIGO, respectively.

\begin{table*}[]
\captionsetup{skip=10pt}
    \caption{{Ablation Study on BlindINDIGO.}}
    \label{tab:as_blind}
    \centering
    \begin{tabular}{|c|c|c|c|c|c|c||c|c|c|}
\hline 
\multirow{5}{*}{\makecell{Synthetic \\ Seen}}&Case  &Sampling&T&N&Initialization 
&Finetuning  & PSNR $\uparrow$& LPIPS $\downarrow$ & Time   \\ \cline{2-10} 
&1&DDPM &1000  &400  &\ding{51}& \ding{51}&25.08&\textbf{0.2334} & 43.85s \\ 
&2&DDIM  &250 &100  &\ding{51}& \ding{51}& 25.28&0.2443 &11.03s \\ 
&3&DDIM   &250 &100  &\ding{51} & \ding{55}&\textbf{25.31} &{0.2442}&9.25s \\
&4& DDIM   &250 &100 & \ding{55}& \ding{55} &24.08  &0.3455&\textbf{9.19s }\\ 
\hline\hline
\multirow{4}{*}{\makecell{Synthetic \\ Unseen}}&Case &Sampling&T&N&Initialization &Finetuning & PSNR $\uparrow$& LPIPS $\downarrow$ & Time   \\ \cline{2-10} 
&5&DDPM  &1000 &400  &\ding{51}& \ding{51}&{24.67} & \textbf{0.2196}& 43.85s\\
&6&DDIM  &250 &100  &\ding{51}& \ding{51}&\textbf{25.18} & {0.2280}&11.03s \\ 
&7&DDIM   &250 &100  &\ding{51} & \ding{55} & 23.19& 0.2546 &\textbf{9.25s} \\

\hline\hline
\multirow{5}{*}{\makecell{Real \\ Unseen}}&Case &Sampling&T&N&Initialization &Finetuning & FID $\downarrow$& NIQE $\downarrow$ & Time   \\
\cline{2-10} 
&8&DDPM&1000 &400  &\ding{51}& \ding{51}  &\textbf{120.42} &{4.2694} & 43.85s \\ 
&9&DDIM&250 &100  &\ding{51}& \ding{51}  &{127.36} &\textbf{4.2297} &11.03s \\ 
&10&DDIM&250  &100 &\ding{51}&\ding{55}  &136.37& 4.6045&\textbf{9.25s}\\  
&11&DDIM&250 &100 &\ding{55}&\ding{51} &155.14 & 5.7365& 10.94s\\ 
\hline  

    \end{tabular}
\end{table*}

\renewcommand{\arraystretch}{1.2} 
\begin{table}[]
    \caption{Ablation Study on the loss function in terms of PSNR$\uparrow$/ LPIPS$\downarrow$ in the case of medium degradation on a subset of CelebAHQ-Test dataset.}
    \label{tab:ablation_inn}
    \centering
    \begin{tabular}{|l||c|c|}
\hline 
             Strategies                            & PSNR&LPIPS   \\ \hline \hline 

No guidance during sampling                   & 24.00 & 0.2685    \\  
$L_{pix,img}$  &24.52 &0.2535    \\  
$L_{pix,y}$ &\textbf{25.24} &0.2452    \\  
$L_{pix,y}+L_{pix,img}$ &{25.21} &  0.2441  \\  
$\mathbf{L_{pix,y}+L_{fea,img}}$(default) & 25.00& \textbf{0.2323}  \\  
$L_{pix,y}+L_{fea,img}+L_{pix,img}$ & 24.95&  0.2330  \\

\hline

    \end{tabular}
\end{table}

\renewcommand{\arraystretch}{1.2} 
\begin{figure}[!tp]\footnotesize 
	\centering
\begin{tabular}{c@{\extracolsep{0em}}c@{\extracolsep{0em}}c@{\extracolsep{0em}}c@{\extracolsep{0em}}c}
   & Low-Resolution &  w/o fine-tuning  &    w/ fine-tuning   \\
          &\includegraphics[width=0.13\textwidth]{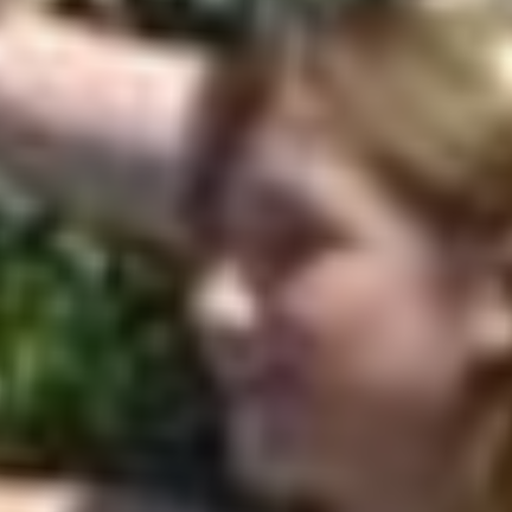}~
        &\includegraphics[width=0.13\textwidth]{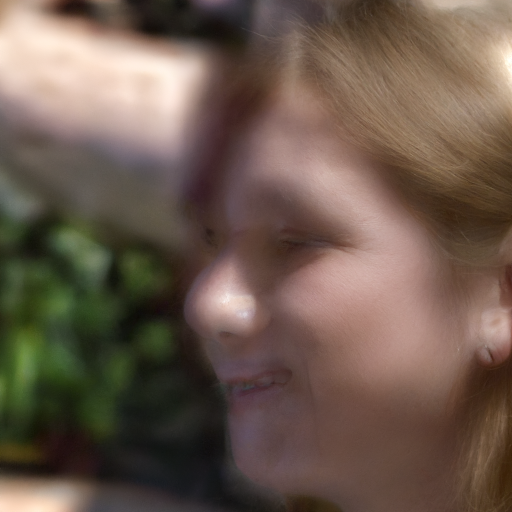}~
               &\includegraphics[width=0.13\textwidth]{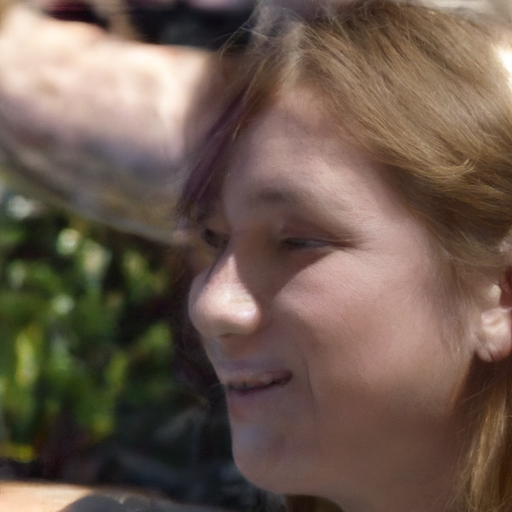}~\\
&\includegraphics[width=0.13\textwidth]{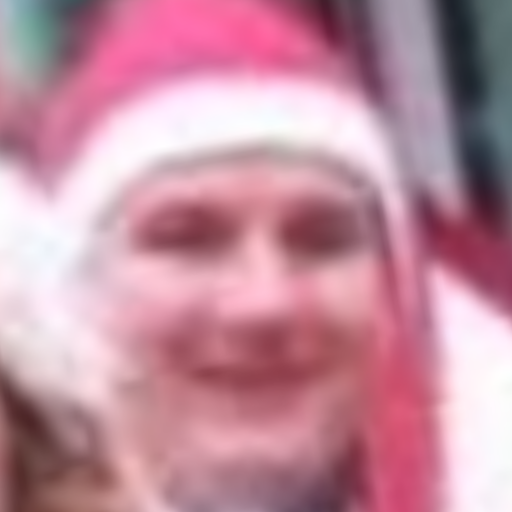}~
        &\includegraphics[width=0.13\textwidth]{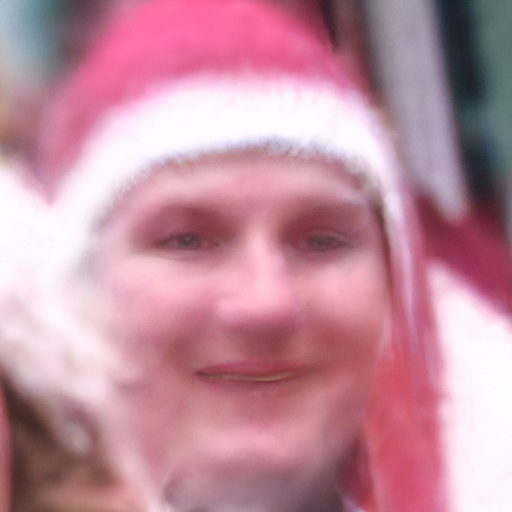}~
               &\includegraphics[width=0.13\textwidth]{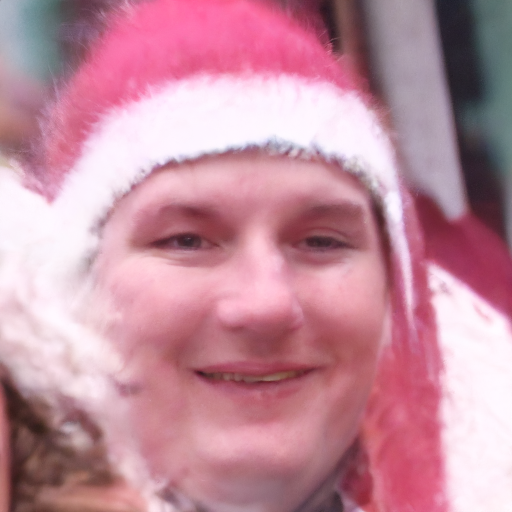}~\\

	\end{tabular}
	\caption{ 
Some examples where our algorithm with the \textit{pre-trained} INN does not perform well under severe and complex real-world degradation conditions (second column). This issue is fixed with our \textit{fine-tuning} strategy (third column).} \vspace{-0.3cm}
	\label{fig:finetune}
\end{figure}

\subsubsection{Effect of loss function}
\label{loss_ablation}
We explore different loss function designs for our BlindINDIGO in Table \ref{tab:ablation_inn}, where $L_{pix,y}$, $L_{pix,img}$, $L_{fea,img}$ denote MSE loss in measurement space between $\bm{c}_{t}$ and $\bm{y}$, MSE loss in image space between $\hat{\bm{x}}_{0,t}$, $\bm{x}_{0,t}$ and perceptual loss in image space between $\hat{\bm{x}}_{0,t}$, $\bm{x}_{0,t}$, respectively. One can observe that the result with loss function $L_{pix,y}$ achieves best PSNR, while the result with loss function $L_{pix,y}$ and $L_{fea,img}$ achieves best perceptual quality. We use the loss function $L_{pix,y}$ and $L_{fea,img}$ in all our experiments for the blind case.

\begin{figure}[!tp]\footnotesize 
	\centering
\begin{tabular}{c@{\extracolsep{0em}}c@{\extracolsep{0em}}c@{\extracolsep{0em}}c@{\extracolsep{0em}}c}
   & Low-Resolution &  T=250  &  T=1000  &    Ground Truth   \\
          &\includegraphics[width=0.12\textwidth]{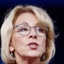}~ &\includegraphics[width=0.12\textwidth]{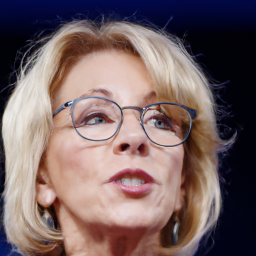}~
        &\includegraphics[width=0.12\textwidth]{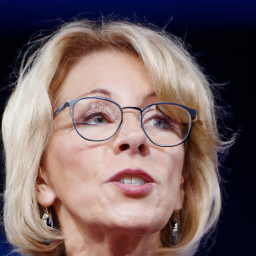}~
               &\includegraphics[width=0.12\textwidth]{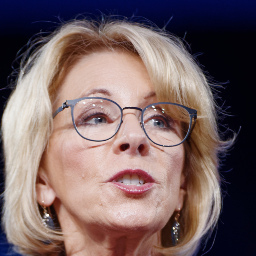}~\\
	\end{tabular}
	\caption{ 
{Results of our approach with T=1000 and T=250 (with DDIM) on 4x super-resolution task.}} \vspace{-0.3cm}
	\label{fig:ddim}
\end{figure}

{
\subsubsection{Sampling With DDIM}
\label{ddim_discuss}
To accelerate the sampling process, we use DDIM \cite{ddim} as the sampling strategy, which skips steps in the reverse process to speed up the DDPM generating process.
We show the performance in terms of PSNR and LPIPS with respect to the change in NFEs (number of neural function evaluations) on the first 100 images of the testing dataset in Table \ref{tab:nfe}.} {One can observe that our algorithm achieves good performance when NFE $>=$250, whereas when NFE=100, performance deteriorates significantly.
Moreover, as it can be seen in Fig. \ref{fig:ddim}, both versions of our algorithm, that is, the default version based on DDPM with T=1000 and the fast version based on DDIM with T=250, produce high-quality reconstruction results while preserving realistic details. In addition, we apply DDIM as the sampling strategy in our BlindINDIGO and present the results in Table \ref{tab:as_blind}. As shown in Table \ref{tab:as_blind}, we investigate effects of different components in our BlindINDIGO. We apply our BlindINDIGO on 3 settings: synthetic seen degradation, synthetic unseen degradation and real unseen degradation. For synthetic seen degradation, we test with medium degradation ($\sigma$=6, $\delta$=25, $q$=50). For synthetic unseen degradation, we set 
$\sigma$=40, $\delta$=0, and $q$=100. Both of them are evaluated on a subset (first 100 images) of CelebA HQ validation dataset. For the real unseen degradation setting, we evaluate our approach on a subset (first 100 images) of WIDER-Test dataset. One can observe the effect of DDIM sampling from cases 1 and 2, from cases 5 and 6, or from cases 8 and 9 in Table \ref{tab:as_blind}.
Overall, this analysis confirms that our approach is compatible with DDIM and that DDIM can be used to accelerate the reconstruction process with only a small compromise to the perceptual quality. 
}

{
\subsubsection{Effect of initialization} As shown in Table \ref{tab:as_blind}, we investigate the effect of the initialization in our BlindINDIGO. 
When comparing cases 3 and 4, or when comparing cases 9 and 11, we can see that with our initialization strategy, our approach performs better in terms of both reconstruction accuracy and image quality.}
\subsubsection{Effect of finetuning} As discussed in Section \ref{blind}, in real-world scenarios with more complex degradations, the parameters of our INN need to be refined to simulate the degradation process more accurately. We achieve this by finetuning the parameters of our INN at testing stage. {We investigate the effect of finetuning in our BlindINDIGO in Table \ref{tab:as_blind}. One can observe that our finetuning strategy contributes to performance improvement in both synthetic unseen degradation (cases 6 and 7) and real-world unseen degradation (cases 9 and 10).} As shown in Fig. \ref{fig:finetune}, without finetuning, the output image becomes blurry due to the inaccurate simulation of the degradation process. Finetuning fixes this issue.

\section{Conclusion}
In this paper, we have introduced a novel approach that fully leverages the power of pre-trained generative diffusion models for inverse problems. We achieve this by introducing an INN that enforces that the generative process of the diffusion model be consistent with the measurements. This leads to a simple way to effectively sample from the posterior rather than the prior as in unconditional diffusion. 

Besides being very effective, the approach is extremely flexible since the degradation process can be learned from data and refined at testing stage if necessary. 
In the non-blind case, since we pre-train the forward process of INN to simulate an arbitrary degradation process, we are no longer limited by the requirement of knowing the analytical expression of the degradation model and we can handle highly non-linear degradation processes. In the blind case, we can handle unknown degradations due to our approach that at testing stage alternately refine the INN to better simulate the unknown degradation and update intermediate results with the guidance of INN during reverse diffusion sampling.

Experiments demonstrate that our algorithm obtains competitive results 
both quantitatively and visually on synthetic and real-world low-quality images.

{\appendices
{\section{Consensus Strategy}
\label{appendix}
We propose a novel consensus strategy for our INDIGO. Our insight is that the Langevin iteration in Eq.~\ref{xt-langevin} has a random term $\bf{z}$. We can therefore create several parallel versions of that iteration by using different realization of $\bf{z}$. In this way our method estimate several enhanced versions of the corrupted image that can then be combined. However, instead of directly averaging the outputs of our algorithm, we adopt an averaging operation during the guidance of the gradient after each sampling step as shown in Algorithm~\ref{algo:blind_repeat}. It is noteworthy that our strategy supports averaging multiple results. However, we show only the case of averaging two results in the algorithm for simplicity.

\begin{algorithm}[htb]
\small
	\caption{{INDIGO with Consensus Strategy}}

	\KwIn{Corrupted image $\boldsymbol{y}$, gradient scale $\zeta$, pretrained INN $f_{\phi}(\cdot)$.}
	\KwOut{Output image $\boldsymbol{x}_{0}$ conditioned on $\boldsymbol{y}$}

    $\bm{x}_T^{1} \sim \mathcal{N}(\bzero, \bm{I})$

    $\bm{x}_T^{2} \sim \mathcal{N}(\bzero, \bm{I})$
    
	\For{$t$ from $T$ to 1}{
        $\bm{z}^{1} \sim \mathcal{N}(\bzero, \bI)$ if $t > 1$, else $\bm{z}^{1} = \bzero$

        $\bm{z}^{2} \sim \mathcal{N}(\bzero, \bI)$ if $t > 1$, else $\bm{z}^{2} = \bzero$
        
        $\bm{x}_{0,t}^{1}  = \frac{1}{\sqrt{\bar\alpha_t}}(\bm{x}_{t}^{1} - \sqrt{1 - \bar\alpha_t} \bepsilon_\theta(\bm{x}_t^{1}, t) )$
    
        $\bm{x}_{0,t}^{2}  = \frac{1}{\sqrt{\bar\alpha_t}}(\bm{x}_{t}^{2} - \sqrt{1 - \bar\alpha_t} \bepsilon_\theta(\bm{x}_t^{2}, t) )$
        
        $\tilde{\bm{x}}_{t-1}^{1}  = \frac{\sqrt{\alpha_t}(1-\bar\alpha_{t-1})}{1 - \bar\alpha_t}\bm{x}_{t}^{1}+\frac{\sqrt{\bar\alpha_{t-1}}\beta_t}{1 - \bar\alpha_t}\bm{x}_{0,t}^{1}  + \sigma_t \bm{z}^{1}$

        $\tilde{\bm{x}}_{t-1}^{2}  = \frac{\sqrt{\alpha_t}(1-\bar\alpha_{t-1})}{1 - \bar\alpha_t}\bm{x}_{t}^{2}+\frac{\sqrt{\bar\alpha_{t-1}}\beta_t}{1 - \bar\alpha_t}\bm{x}_{0,t}^{2}  + \sigma_t \bm{z}^{2}$
        
        {\color{blue}{$\bm{c}_{t}^{1},\bm{d}_{t}^{1} =f_{\phi}(\bm{x}_{0,t}^{1})$}}

        {\color{blue}{$\bm{c}_{t}^{2},\bm{d}_{t}^{2} =f_{\phi}(\bm{x}_{0,t}^{2})$}}
        
        {\color{blue}{$\hat{\bm{x}}_{0,t}^{1} =f_{\phi}^{-1}(\bm{y},\bm{d}_{t}^{1})$}}

        {\color{blue}{$\hat{\bm{x}}_{0,t}^{2} =f_{\phi}^{-1}(\bm{y},\bm{d}_{t}^{2})$}}

        {\color{blue}{$\bm{x}_{t-1}^{1} =\tilde{\bm{x}}_{t-1}^{1}  - {{\zeta}}\nabla_{\bm{x}_{t}^{1}} ( \|{\hat{\bm{x}}_{0,t}^{1}- \bm{x}_{0,t}^{1}}\|_2^2+\|{\hat{\bm{x}}_{0,t}^{2}- \bm{x}_{0,t}^{1}}\|_2^2)$}} 

        {\color{blue}{$\bm{x}_{t-1}^{2} =\tilde{\bm{x}}_{t-1}^{2}  - { {\zeta}}\nabla_{\bm{x}_{t}^{2}}(\|{\hat{\bm{x}}_{0,t}^{2}- \bm{x}_{0,t}^{2}}\|_2^2+\|{\hat{\bm{x}}_{0,t}^{1}- \bm{x}_{0,t}^{2}}\|_2^2)$}}

	}
	\Return $\boldsymbol{x}_{0}$
\label{algo:blind_repeat}
\vspace{-0.1em}
\end{algorithm}
\begin{table}[htb]
\small
    \caption{Ablation study on our consensus strategy on non-blind 4x super-resolution.}
    \label{tab:consensus}
    \centering
    \begin{tabular}{|c|c|c|c|}
\hline 
            Noise level& Strategies                            & PSNR  &PSNR gain \\ \hline \hline

\multirow{4}{*}{{30}} &Baseline                       & 23.97 & --            \\  
&Averaging 2 results                       & 24.12& 0.15          \\  
&Averaging 3 results                      &\textbf{24.51}& \textbf{0.54}     \\
&Averaging 4 results                   & 24.34& 0.37          \\
\hline  \hline 
\multirow{4}{*}{50} &Baseline                       & 22.35 & --            \\  
&Averaging 2 results                       & 22.56& 0.21          \\  
&Averaging 3 results                      & \textbf{23.03}& \textbf{0.68}      \\
&Averaging 4 results                   & 22.91& 0.56          \\
\hline  \hline 
\multirow{4}{*}{80} &Baseline                       & 20.51 &--            \\  
&Averaging 2 results                       & 20.91& 0.40          \\  
&Averaging 3 results                      & \textbf{21.50}& \textbf{0.99}      \\
&Averaging 4 results                   & 21.48& 0.97          \\
\hline

    \end{tabular}
\end{table}

\begin{table}[]
\captionsetup{skip=10pt}
    \caption{{Ablation study on the consensus strategy of our BlindINDIGO on 4x blind SR with medium degradation.}}
    \label{tab:as_blind_consensus}
    \centering
    \begin{tabular}{|c||c|c|c|}
\hline 
  & PSNR $\uparrow$& LPIPS $\downarrow$ & Time   \\ \hline
w/o Consensus & \textbf{25.31} &{0.2442}&\textbf{9.25s} \\
w/ Consensus &25.26 &\textbf{0.2413} &19.98s \\

\hline  
 
    \end{tabular}
\end{table}

In Table \ref{tab:consensus}, we show our INDIGO with up to four parallel versions with our consensus strategy in the non-blind case. It can be seen that our consensus strategy with 3 parallel versions achieves the best performance. Also, we can observe that as the noise level increases, the PSNR gain brought by our strategy is more significant. Therefore, in the non-blind case, we apply our consensus strategy with 3 parallel results. {In the blind case, as shown in Table \ref{tab:as_blind_consensus}, we observe that although our consensus strategy brings gains in terms of LPIPS, it also has an impact on the runtime. Taking into account the tradeoff between improved image quality and the additional computational time required, we decide not to employ our consensus strategy in the blind case.}}}

\bibliographystyle{IEEEtran}
\bibliography{ref}
\newpage

\begin{IEEEbiography}[{\includegraphics[width=1in,height=1.25in,clip,keepaspectratio]{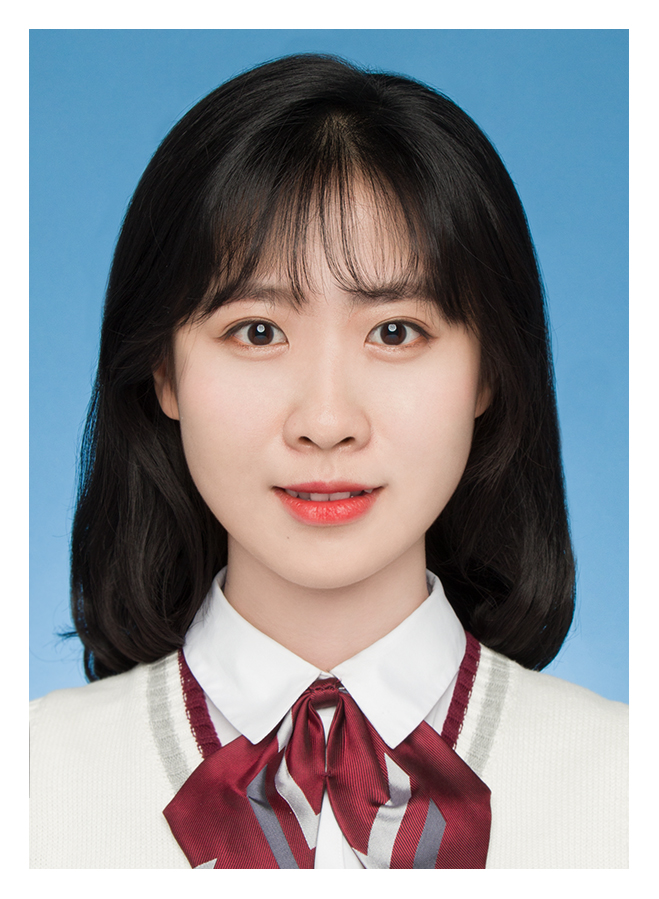}}]{Di You} (Student Member, IEEE) received the bachelor’s degree in electronic and information engineering from Dalian University of Technology, Dalian, China, in 2019, and the master’s degree in computer applications technology from Peking University, Shenzhen, China, in 2022. She is currently working toward the Doctoral degree with Electrical and Electronic Engineering Department, Imperial College London, London, U.K., under the supervision of Professor Pier Luigi Dragotti. Her research interests include the areas of computer vision, signal processing, and deep learning, specifically for inverse problems. She was awarded the President’s Ph.D. Scholarship by Imperial College London.
\end{IEEEbiography}
\vspace{-250pt}
\begin{IEEEbiography}[{\includegraphics[width=1in,height=1.25in,clip,keepaspectratio]{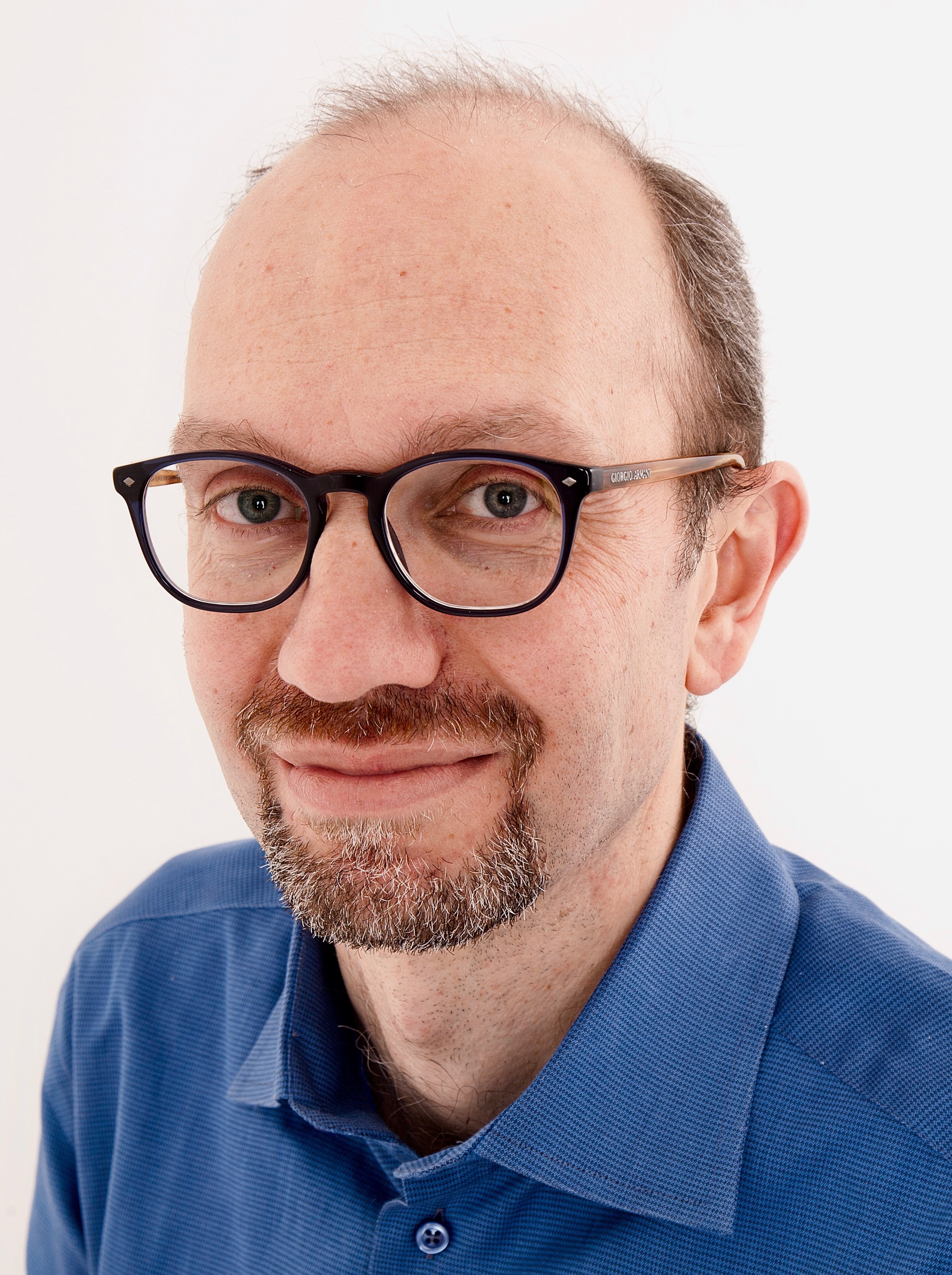}}]{Pier Luigi Dragotti}
(Fellow, IEEE) received the Laurea degree (\textit{summa cum laude}) in electronic engineering from the University of Naples Federico II, Naples, Italy, in 1997, and the master’s degree in communications systems and the Ph.D. degree from the Swiss Federal Institute of Technology of Lausanne (EPFL), Switzerland, in 1998 and 2002, respectively. He has held several visiting positions, in particular, he was a Visiting Student with Stanford University, Stanford, CA, USA, in 1996, a Summer Researcher with the Mathematics of Communications Department, Bell Labs, Murray Hill, NJ, USA, in 2000, a Visiting Scientist with the Massachusetts Institute of Technology, Cambridge, MA, USA, in 2011, and a Visiting Scholar with Trinity College, Cambridge, U.K., in 2020. Before joining Imperial College London, London, U.K., in November 2002, he was a Senior Researcher with EPFL working on distributed signal processing for the Swiss National Competence Center in Research on Mobile Information and Communication Systems. He is currently a Professor of signal processing with the Department of Electrical and Electronic Engineering, Imperial College London. His research interests include sampling theory and its applications,
computational imaging, and model-based deep learning. Dr. Dragotti was an Elected Member of the IEEE Image, Video and Multidimensional Signal Processing Technical Committee as well as an Elected Member of the IEEE Signal Processing Theory and Methods Technical Committee and the IEEE Computational Imaging Technical Committee. In 2011, he was the recipient of the Prestigious ERC Starting Investigator Award (consolidator stream). He was also IEEE SPS Distinguished Lecturer (2021–2022), Editor-in-Chief of the IEEE TRANSACTIONS ON SIGNAL PROCESSING (2018–2020),
Technical Co-Chair of the European Signal Processing Conference in 2012 and
an Associate Editor for IEEE TRANSACTIONS ON IMAGE PROCESSING from 2006 to 2009.
\end{IEEEbiography}
\vspace{11pt}

\end{document}